\documentclass[journal]{IEEEtran}
\usepackage[utf8]{inputenc}
\usepackage[T1]{fontenc}
\usepackage{algpseudocode}
\usepackage{algorithm}
\usepackage{amsmath}
\usepackage{upgreek}
\usepackage{amssymb}
\usepackage{graphicx}
\usepackage{stfloats}
\usepackage{tabularx}
\usepackage{rotating}
\usepackage[dvipsnames,table]{xcolor}

\ifCLASSINFOpdf
\else
\fi
\begin{document}
%
\title{Consensus of state of the art mortality prediction models: From all-cause mortality to sudden death prediction}
%
%
%

\author{Yola Jones, Dr Fani Deligianni, Dr Jeff Dalton, Dr Pierpaolo Pellicori, Professor John G F Cleland}

%
%

\markboth{Journal of \LaTeX\ Class Files,~Vol.~14, No.~8, August~2015}%
{Shell \MakeLowercase{\textit{et al.}}: Bare Demo of IEEEtran.cls for IEEE Journals}
%



\maketitle

\begin{abstract}
Worldwide, many millions of people die suddenly and unexpectedly each year, either with or without a prior history of cardiovascular disease. Such events are sparse (once in a lifetime), many victims will not have had prior investigations for cardiac disease and many different definitions of sudden death exist. Accordingly, sudden death is hard to predict.

This analysis used NHS Electronic Health Records (EHRs) for people aged $\geq$50 years living in the Greater Glasgow and Clyde (GG\&C) region in 2010 (n = 380,000) to try to overcome these challenges. We investigated whether medical history, blood tests, prescription of medicines, and  hospitalisations might, in combination, predict a heightened risk of sudden death.

We compared the performance of models trained to predict either sudden death or all-cause mortality. We built six models for each outcome of interest: three taken from state-of-the-art research (BEHRT, Deepr and Deep Patient), and three of our own creation. We trained these using two different data representations: a language-based representation, and a sparse temporal matrix.

We used global interpretability to understand the most important features of each model, and compare how much agreement there was amongst models using Rank Biased Overlap. It is challenging to account for correlated variables without increasing the complexity of the interpretability technique. We overcame this by clustering features into groups and comparing the most important groups for each model. We found the agreement between models to be much higher when accounting for correlated variables.

Our analysis emphasises the challenge of predicting sudden death and emphasises the need for better understanding and interpretation of machine learning models applied to healthcare applications.

\end{abstract}


%
\IEEEpeerreviewmaketitle

\section{Introduction}
Worldwide each year, millions of people die suddenly and unexpectedly \cite{RN311,RN355,RN358,RN361,RN369,RN291,RN297,RN294}. Sudden death (SD) often occurs in the setting of chronic disease long before the condition has become terminal but may also be the first and only manifestation of disease, most often cardiovascular.  For many individuals, SD is not preceded by any warning symptoms that could prompt clinical investigations or an admission to hospital. Current estimates of the incidence of SD vary widely because of heterogeneity in cohort inclusion criteria and how SD is defined. Moreover, in clinical practice, SD occurring out of hospital is commonly certified as due to myocardial infarction. Probably only a minority of SD are reported as such. Accordingly, it is not surprising that predicting SD is challenging. 

The availability of a large volume of routinely collected, longitudinal, electronic health records (EHR), including blood results, prescriptions, imaging investigations and diagnoses provides opportunities to try to predict SD, which might offer insights into underlying mechanisms and potential prevention or management strategies. However, EHRs are complex including thousands of clinical variables and many data types collected at different times and for different reasons. Without prior expert knowledge, conventional statistical models cannot cope. Despite the recent increase in the use of machine learning in healthcare, developing and deploying deep learning for risk prediction with EHRs is still an emerging field. The nature of Electronic Health Records (EHRs) accumulated for over a decade add further challenges, since the risk prediction relies on recognising long-term dependencies between sparse, irregular events such as hospitalisations, lab tests and combinations of prescriptions.

Steyerberg et al. highlighted the importance of clinical utility and probability calibration for evaluating clinical prediction models \cite{RN277}. Although, in 2014 Steyerberg et. al set guidelines for models development and validation \cite{10.1093/eurheartj/ehu207}, but only recenrlt have these recommendations been extended to risk prediction models using artificial intelligence and machine learning \cite{RN277,10.1093/eurheartj/ehac238}.  
Hond et. al \cite{RN277} highlighted interpretability and model transparency as prerequisites for healthcare applications \cite{RN192,RN195}. Interpretability highlights the importance of input features and helps identify model biases inherent in healthcare datasets \cite{RN196,RN197,RN207,RN208}. This should enhance clinicians' trust in machine learning by providing them with a model they can understand, and allow extraction of useful insights from complex data.  

In this paper, we use EHRs to adapt and compare five state-of-the art learning models along with a baseline logistic regression model for the prediction of SD and other catastrophic cardiovascular (CV) events (myocardial infarction (MI), stroke, survived cardiac arrest), and all-cause mortality in people aged 50 years, or older, served by NHS Greater Glasgow and Clyde, the largest health board and healthcare provider in Scotland. Three of these models are taken from existing state-of-the-art open-source models for mining EHR \cite{RN261,RN264,RN265}, whereas others are inspired from research in deep learning and have been developed in-house. These six models cover two different data representations: a sparse temporal matrix, and a language representation. Our population of around 380,000 are people aged 50 or older as of the 1st of January 2010, with a Greater Glasgow and Clyde postcode.  In order to understand better the performance of these models, we also employ them in predicting all-cause mortality. We find that all models perform better when predicting all-cause mortality (AUC of 0.85) versus sudden death (AUC of 0.77). All-cause mortality includes patients with life-shortening conditions, such as cancer and heart failure, whose shortened survival is fairly predictable. Predicting sudden death in people who appear healthy, or at least fairly well, is more difficult.

Furthermore, we investigated global interpretability to understand which features have the greatest impact on  performance. We found that relying on single feature permutation strategies, as reported in previous analysis \cite{9706318}, is not appropriate to estimate the significance of input features, since clinical variables are highly inter-correlated. This also compromises the agreement between models in terms of feature ranking. To address this problem we exploit spectral-coclustering, which clusters patients and features simultaneously into mutually exclusive features using spectral graph partitioning. We find agreement between models to improve when clusters of features are taken into account from 0.36 to 0.66 and from 0.32 to 0.71 Rank Biased Overlap (RBO) for all cause-mortality and sudden death prediction, respectively.

 \section{Background}
\subsection{Deep Prediction Models based on EHR}
The study by Nguyen et. al in 2016 was one of the first that enabled automatic feature representation derived from EHRs without extensive knowledge of domain experts \cite{RN264}. Feature representations based on bag-of-words could not encode appropriately the temporal relationships present in EHRs. For this reason, they created DeepR, a language representation model for EHRs using convolutional neural networks (CNNs). DeepR is designed to transform an EHR of a patient into a 'sentence' of 'multiple phrases', which encode visit episodes. These phrases are separated by 'special words', which represent time gaps, tracking the evolution of patient's health over time. Sentences are passed via the CNN layers to form a global feature vector of local associations in a fully-supervised fashion to predict risk of future events. The authors trained this model using hospitalisation data from 300K patients from a hospital chain in Australia. In 2020, Roberto, J. et al \cite{RN262} expanded this model to include demographic information like age and sex.

Also in 2016, Miotto et al. proposed Deep Patient \cite{RN265}, which encodes patient's records representation with an unsupervised learning model consisting of a stack of denoising autoencoders which use sparse vectors to represent patient records as an input. Using this representation, each feature is a different type of event (e.g. prescription of medications, diagnosis codes), with the entry at each being the count of each time the event occurred. This model allows unlabelled data to be used to pre-train a model, which can then be fine-tuned for a variety of downstream tasks. 

More recently, in 2020, Li et. al proposed BEHRT \cite{RN261}, a language-based healthcare model taking inspiration from the hugely successful language model BERT \cite{BERT_ref}. Like BERT, BEHRT uses a transformer architecture for unsupervised pre-training in a particular domain, in this case for use with EHRs. With this architecture, unlabelled patient records are represented as sentences and are used to pre-train a classifier using a Masked Language Model (MLM), which is fine tuned for a variety of downstream tasks.

For this study, we build three of the models discussed above, namely BEHRT, Deepr, and Deep Patient. We train each of these on the NHS Greater Glasgow and Clyde Safe Haven dataset, for the prediction of SD and other catastrophic CV events.

\subsection{Interpretability in Machine Learning for Healthcare}
Both algorithmic transparency and explainability have been highlighted as a fundamental prerequisite in deploying machine learning models in healthcare \cite{RN412,RN414,RN413}.  However, what constitutes an adequate explanation is obscure. In its very basic form, explainability summarises relationships contained in data or learned by the model in such a way that they reveal the importance of the input feature to the model's decision \cite{RN414}. At a 'global' level these methods resemble traditional feature importance and variable selection across the whole dataset whereas in a 'local' level they operate within the neighbour of a specific decision \cite{RN263}. 

In high dimensional spaces with hundreds of input variables, feature importance needs to be coupled with knowledge of the inter-relationships between these variables. If clinical variables are correlated, they constitute clusters of inter-changeable information. This is an important factor to consider when interpreting the decision of model and handling missing information. 

 \subsection{Biclustering}
 Biclustering is an unsupervised learning technique used to group data into clusters using multiple dimensions (rows and columns), unlike traditional clustering methods which only cluster rows. It is typically used for clustering genetic data when both genes and associated conditions are needed to obtain meaningful clusters \cite{biclustering_genes}, in which it is important for data points to be able to be clustered into more than one group. This technique has been used by Divina et al \cite{RN539} who used biclustering to analyse the energy consuption of homes, with the aim of clustering both by location and time to find buildings which use similar amounts of energy during particular time periods. More recently, Dhamodharavadhani and Rathipriya used this techique to find countries with a similar epidemic trajectory using the COVID-19 pandemic across specific time periods, used to inform public health in future \cite{Dhamodharavadhani2021}.

 In this study, we use Biclustering to cluster both features and patient simultaneously,to compare the similarity between features with respect to specific patient groups.

\section{Methods}
\subsection{Dataset Description and Outcome Definitions}
Pseudo-anonymised, administrative healthcare records (“SafeHaven”) routinely collected by the NHS Greater Glasgow and Clyde (GG\&C) health board between the 1st of January 2010 and the 1st of January 2022 were obtained using the “Glasgow SafeHaven” trusted research environment for people aged 50 years or older as of the 1st of January 2010 with a GG\&C postcode \cite{RN100}. We chose to focus on older people as the risk of SD and all-cause mortality rise steeply with age. We obtained five sets of information including demographic information (age and sex), medical history of major diseases, blood tests, prescriptions, and hospitalisation records \cite{RN264,RN265,RN247,RN249,RN261}.

Our primary outcome of interest was SD, defined as death which occurs out-of-hospital or within one day of a hospital admission, in those who have no evidence of a terminal illness in their medical record in the year before death. Since catastrophic cardiovascular events such as stroke, myocardial infarction, and near-fatal arrhythmia are often considered a 'failed' attempt at sudden death, we also include these events in our definition. Therefore, our definition of sudden death includes both terminal events (an unexpected death) and non-terminal events (survival of catastrophic cardiovascular events). We compare the prediction of sudden death (including other catastrophic cardiovascular events) to the prediction of all-cause mortality. All-cause mortality is defined to be a death for any reason irrespective of prior medical history. More details about the target outcome can be found in Appendix A.

\subsection{Data Representation}

Both the sparsity and the irregular sampling of EHR challenges the development of robust machine learning models. Here, we adapt two representations: a language-based interpretation in which healthcare records are encoded as sentences (as used in BEHRT \cite{RN261} and Deepr \cite{RN264}), and a sparse matrix representation (as used in Deep Patient \cite{RN265}).

Both BEHRT and Deepr use language models with integer encoding to represent EHRs. This representation has the advantage of being information dense, and relatively simple for a human to understand once they are familiar with the encoding system. These language representations result in unequal sequence lengths and a large vocabulary (e.g. ICD-10 codes which have over 68,000 possible codes).

However, contrary to language data, EHRs are pre-processed during the data cleaning and collation process which is required in healthcare datasets. This typically involves correcting sources of noise (for example correcting typos), removing or correcting misentered data, converting coding standards to make them compatible with other datasets, and censoring data for patients who move out of the region. Converting records to sentences requires mapping tabular data to processed free-text, which might remove valuable information. Details on how we clean the data included in this analysis can be found in Appendix B.

Alternatively, Deep Patient uses a sparse matrix encoding, in which patient history is encoded as a matrix of dimensions \textit{features x time}, where time is split into discrete elements. This creates a sparse 2D array for each patient record, of fixed size. This representation is harder for a human to read, but does not require any extra pre-processing, as each patient record is already encoded as a sparse numerical matrix.

A comparison between these two representations is shown in Figure \ref{fig:ehr_to_model_diagram}, and in Appendix C. In order to compare interpretations across different models, we need to standardise and harmonise representations. We do this by using a bag-of-words language embedding for our language models, which turns an input sentence into a sparse vector of size \textit{n\_features}, allowing easier comparison between our sparse matrix and language models.

\begin{figure}[!t]
\centering
\includegraphics[width=0.48\textwidth]{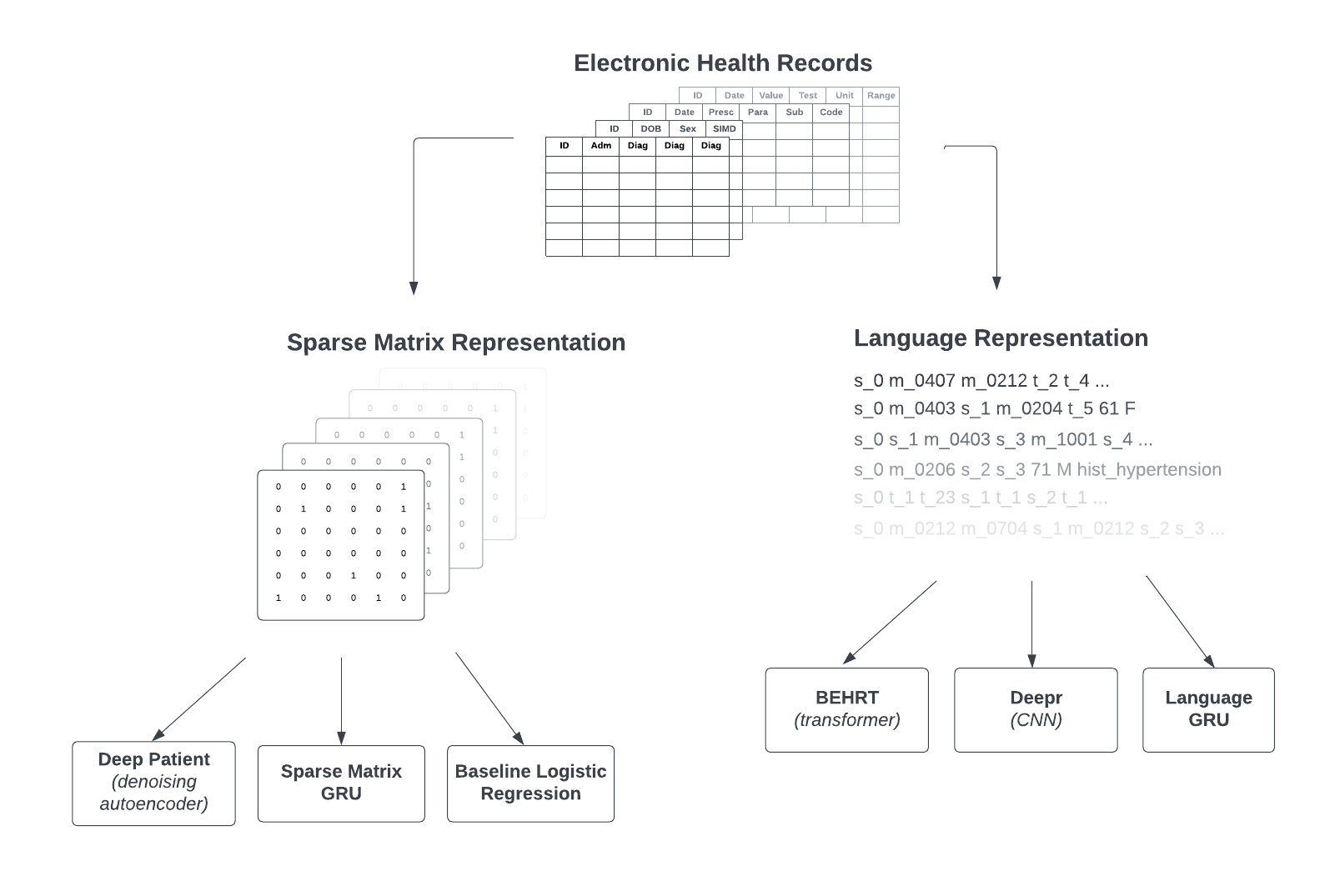}
\caption{Electronic Health Records to Model Diagram \textmd{This figure shows how we take electronic health records and pass them into six different machine learning models.}}
\label{fig:ehr_to_model_diagram}
\end{figure}

To allow interpretations from these representations to be compared, we use the same set of features for both representations. We represent each feature as a word in the language representation, and as a cell value in the sparse matrix representation. We focus on five families of features: prescriptions, blood tests, hospitalisations, demographics, and history of major disease. As mentioned, we use the same set of features for both the language representation and the sparse temporal matrix representation. We describe how we process each feature family and create each data representation in Appendix D.

\subsection{Models Construction}

We include six different models in this study which use two different representations: three language models and three sparse matrix representations. For both of these representations, and to overcome the challenges of longitudinal healthcare datasets, bin each patients medical history into two-month intervals, and count events of each feature within each bin. We do this for the year prior to index, with goal of predicting adverse events within the following six months.

Our analysis involves three state-of-the-art models, namely Deepr (Nguyen, P. et al. \cite{RN264}), Deep Patient (Miotto, R. et al. \cite{RN265}) and BEHRT (Li, Y. et al. \cite{RN261}). We also construct two GRU models, one using a language representation and one using a sparse temporal matrix. We compare all five of these to a baseline logistic regression model. For the state-of-art-models, we follow the available information about the architecture of each model as published by the original researchers. For parameters which are not explicitly defined in the original research, or for parameters which need adapted for the dataset used in this study, we have tuned these hyperparameters for optimal performance when predicting sudden death (including survived sudden death) as well as all-cause mortality.

We provide a description of how we construct each model in Appendix E.

\subsubsection{Model Pre-Training}
Both BEHRT and Deep Patient include a pre-training step in their model training, which helps the models learn the general structure and key properties of the dataset under test, before being fine-tuned for the outcome of interest. For BEHRT, the model is pre-trained on the entire Safe Haven dataset, irrespective of patient outcomes or prior medical history. To do this, we build a language representation of the medical history of every person in this dataset. From this data, BEHRT is pre-trained using a Masked Language Model (MLM).

For the Deep Patient model, Miotto, R. et al. \cite{RN265} added Gaussian noise to their input data and used a stack of denoising autoencoders to build a model to reconstruct the original input. However, since a large proportion of the variables are binary labels (whether or not an event occurred), adding Gaussian noise would create new data that is unrealistic. Because of this, we instead use a Gaussian distribution to randomly mask parts of the input data at a corruption level of 5\%. In other words, 5\% of the input data is masked. Similar to the original paper, we train a stack of denoising autoencoders to reconstruct the original input.

\subsubsection{Vocabulary Construction}
The vocabulary used by all of the models we used is the set of all unique features in our training dataset, as required by our language representation. We create a separate vocabulary for our sudden death population and our all-cause mortality population, as there are many features in our all-cause mortality population which are indicators of terminal disease and they are not included in the sudden death population. This vocabulary is different to the models used by BEHRT and Deepr, as they contain different variables for prediction of events. For the language models, a feature for unknown words is added as (\textit{'[UNK]'}).

\subsection{Ablation Study}
Finally, to test the features used to train each of the models used in this study, we perform an ablation study across features. We split our 631 features into six feature families: demographics, diagnoses, prescriptions, history of major disease indicators, blood test indicators, and blood test value variables. We train a sparse GRU on different subsets of features from these feature families, and report sensitivity, specificity, AUC and F1 score across each subset. 

We first train six models on each of the six feature families to compare which feature family has the biggest impact to model performance. We then combine feature families to see how the model compares when a mix of features are used. Roberto et al \cite{RN262} compared a variety of different state-of-the art models (RETAIN, Deepr, Deep Patient and RETAIN) across two different feature sets: diagnoses and demographics (DD) and demographics, diagnoses and medications (DDM) for the prediction of emergency readmission and heart failure diagnosis on the CPRD dataset. We use the same combination of feature families in this study. For the diagnosis family, we add history of major disease, as these are categorised from hospital records. The complete list of all ten feature subsets we test is as follows:

\begin{enumerate}
    \item Demographics alone
    \item Diagnoses alone
    \item Prescriptions alone
    \item History of major disease variables alone (categorised from hospital records)
    \item Blood test indicator variables alone
    \item Blood test value indicators alone
    \item Demographics and diagnoses
    \item Demographics, diagnoses, and history of major disease
    \item Demographics, diagnoses, history of major disease, and prescriptions
    \item Demographics, diagnoses, history of major disease, prescriptions and all blood test variables (the complete feature set)
\end{enumerate}

\subsection{Model Interpretations}
We have three different types of representation: a sparse matrix, a language model with bag-of-words encoding, and a language model with integer encoding, shown in Figure \ref{fig:ehr_to_model_diagram}. For both the sparse temporal matrix and the bag-of-words language model, we employ Permutation Feature Importance to extract global interpretability across the event domain, which calculates feature importances by randomly perturbing features and measuring the change in model predictions, similar to existing work \cite{9706318}. For our integer encoding, we extract global interpretations from local interpretations using Local interpretable model-agnostic explanations (LIME, by applying LIME to each sample in this dataset, collecting feature importances for each word in the vocabulary, and averaging feature importance indices for each word across each patient record to extract global interpretations.

The result is a list of global feature importance indices, where each feature is an event (e.g. a medication prescription, a hospital diagnosis, etc), equivalent to the result of global interpretability using Permutation Feature Importance.

For integer encoding models (namely BEHRT and Deepr), we collect local interpretations of every sample in the dataset using LIME (described in detail in Appendix F), and average feature importance indices across events.

We now want to see how much these models agree with each other, by comparing the feature importance rankings. We do this in two ways: qualitatively and quantitatively. We compare these qualitatively by comparing top 10 feature importance indices for each model, and grouping these into distinct feature sets. We then look at what 'groups' of features each model is most interested in. We use quantitative analysis to statistically compare feature importance agreement between models. We do this using Rank-Biased Overlap (RBO) which we explain in detail in Appendix G.

\subsubsection{Clustered Rank Biased Overlap}
Many of the features included in this study are highly correlated and encode similar information, as is true of many healthcare datasets. Both the Rank-Biased Overlap algorithm and the interpretability methods we use in this paper do not consider correlation between features common in healthcare datasets, which does impact the numerical agreement between model feature importance indices. To overcome this, we cluster features into groups, and compare similarity between cluster importance indices for each model. To do this, we first remove the time domain by summing all features across time. For 'static' features which effectively do not change over the course of a year (for example age, sex, and history of major disease), we take the last recorded sample for each patient.

We are then left with a matrix of size $(patients,features)$. We want to compress all patients into a lower dimensional space, in order to capture how similar features are to each other, which we do using biclustering to cluster the dataset used in this study in both patient and feature dimensions simultaneously. This allows us to extract clusters across both the feature domain (in order to find similar features) and patient domain (allowing us to extract patient sub-groups). However, unlike genetic data, our application means we need each data point to be clustered into only one cluster. Because of this, we adopt a biclustering technique called Spectral Co-Clustering, originally proposed by Dhillon in 2001 \cite{coclustering_original}, which allows us to cluster data points into mutually exclusive clusters across two dimensions. Spectral co-clustering is defined by the following two equations \cite{coclustering_original}:

\begin{align}
\mathcal{F}_m=\left\{f_i: \sum_{j \in \mathcal{P}_m} A_{i j} \geq \sum_{j \in \mathcal{P}_t} A_{i j}, \forall l=1, \ldots, k\right\}\\
\mathcal{P}_m=\left\{p_j: \sum_{i \in \mathcal{F}_m} A_{i j} \geq \sum_{i \in \mathcal{F}_l} A_{i j}, \forall l=1, \ldots, k\right\}
\end{align}

where:
\begin{itemize}
    \item $A$ is a n-by-m array, with n representing patients and m representing features
    \item $F$ and $P$ are the interconnected sets of clusters across 'features' and 'patients' dimensions, respectively.
\end{itemize}

Since these two equations depend on each other, these two processes are recursively applied to $A$ until a stable optimum is reached.

Because our patient dimension is much larger than our feature dimension, these clustering algorithms tend to cluster each feature into a single cluster. Therefore, when choosing the optimum number of clusters for this analysis, we attempt to minimise two properties:

\begin{itemize}
    \item The distance between datapoints and their cluster centroids using the Elbown method
    \item The number of clusters which contain only a single feature
\end{itemize}

From here, we extract both feature-level clusters and patient-level clusters. We compare feature-level clusters to the correlation between variables found using the Pearson correlation coefficient, and pass groups of features into our Rank Biased Overlap algorithm to determine the new agreement between features when consider correlated variables. Finally, we analyse patient clusters by looking at distribution of events within each cluster.

\section{Results}

Our final feature set is made up of 629 variables, described in Appendix H. Our sparse matrix has a sparsity of 98.9\%, with each patient being represented by a \textit{629 x 7} temporal matrix, where 7 is the number of time steps (1 year split into 60-day intervals). Our sentence dataset has a maximum sentence length of 209 words, a minimum of 9 words, and a mean of 90.0 words for each person in the dataset (standard deviation 70.8).

\subsection{Prediction of Sudden Death versus All-Cause Mortality}
The Safe Haven includes multi-provider, sparse, longitudinal, and observational data with irregular time points. Most of this population are 'healthy' individuals with no events of interest in the period of observation. Overall, 69,594 met out definition of sudden death (including unexpected death and survived catastrophic cardiovascular events) (median age 78, interquartile range (69, 86), 57\% of whom are women), and 168,434 people in our control group. For all-cause mortality, no exclusion criteria were applied and deaths due terminal illness were included; 108,874 people died (median age 81, interquartile range (73, 87), 53\% of whom are women), with 252,446 people in our control group.

The pre-training training curves for models which use pre-training are given in Appendix I. For the rest of the models (including fine-tuning for BEHRT and Deep Patient), we use an Adam optimizer with a learning rate of 0.001 and a binary crossentropy loss function. We train each model for 10 epochs. The training curves for each of the 5 models can be found in Appendix I.

We report the sensitivity, specificity, F1 score and AUC for all five of the models used in this study, plus our baseline model for the prediction of sudden death (including other catastrophic cardiovascular events) in Table \ref{table:machine_learning_results_sudden_death}. We can see from this table that BEHRT outperforms all other models across all metrics for the prediction of sudden death, achieving an AUC of 0.8279 compared to the next-best model (the sparse GRU) with an AUC of 0.7665. From this table we can see that for all models, sensitivity is much higher than specificity, indicating the models built in this study are better at recognising those that are at risk compared to those that are not at risk.

\begin{table}
\centering
\footnotesize
\begin{tabular}{|p{0.8cm} | p{0.8cm} | p{0.8cm} | p{0.8cm} | p{0.8cm} | p{0.8cm} | p{0.8cm} |} 
 \hline
 \textbf{Metric} & \textbf{Deep Patient (sparse)} & \textbf{Deepr (language)} & \textbf{GRU (language)} & \textbf{GRU (sparse)} & \textbf{BEHRT} & \textbf{LR} \\ [0.5ex]
 \hline
 Sensit - ivity & 0.9255 $\pm$ 0.0145 & 0.8917 $\pm$ 0.0128 & 0.9110 $\pm$ 0.0143 & 0.9253 $\pm$ 0.0083 & \textbf{0.9659 $\pm$ 0.0029} & 0.9325 $\pm$ 0.0005 \\
 \hline
 Specif - icity & 0.4916 $\pm$ 0.0335 & 0.5750 $\pm$ 0.0213 & 0.5845 $\pm$ 0.0291 & 0.6077 $\pm$ 0.0214 & \textbf{0.6898 $\pm$ 0.0134} & 0.5584 $\pm$ 0.0011 \\
 \hline
 F1 score & 0.7270 $\pm$ 0.0075 & 0.7443 $\pm$ 0.0028 & 0.7619 $\pm$ 0.0039 & 0.7830 $\pm$ 0.0042 & \textbf{0.8464 $\pm$ 0.0015} & 0.7652 $\pm$ 0.0002 \\
 \hline
 AUC ROC & 0.7085 $\pm$ 0.0098 & 0.7333 $\pm$ 0.0049 & 0.7477 $\pm$ 0.0076 & 0.7665 $\pm$ 0.0066 & \textbf{0.8279 $\pm$ 0.0053} & 0.7455 $\pm$ 0.0003 \\
 \hline
\end{tabular}
\newline
\caption{Machine Learning Results (sudden death) \textmd{This table gives the precision, recall, sensitivity and specificity of each of our 5 models and our baseline logistic regression model for the prediction of sudden death (including other catastrophic cardiovascular events). Results are given as \textit{mean $\pm$ standard deviation} with each model being run 10 times using K-fold cross validation. The best result for each metric is given in bold.}}
\label{table:machine_learning_results_sudden_death}
\end{table}

For all-cause mortality, the sparse GRU outperforms other models for the prediction of all-cause mortality in all metrics except sensitivity (Table \ref{table:machine_learning_results_all_cause_mortality}). As for models predicting sudden death, all models have a higher sensitivity compared to specificity, indicating that they are better at identifying those that are not at risk of all-cause mortality versus identifying those that are at risk.

Comparing the sudden death prediction results from Table \ref{table:machine_learning_results_sudden_death} to the all-cause mortality results in Table \ref{table:machine_learning_results_all_cause_mortality}, we can see that sudden death is somewhat harder for these models to predict  compared to all-cause mortality, achieving a maximum AUC of 0.8279 (BEHRT) versus 0.8460 (sparse GRU) for all-cause mortality. This is expected, as by definition, the sudden death population excludes terminally ill patients who are expected to die, which are included in the all-cause mortality population. Overall, the sudden death models achieve a mean AUC of 0.7549, compared to the all-cause mortality models which achieve an AUC of 0.8281.

\begin{table}
\centering
\begin{tabular}{|p{0.8cm} |  p{0.8cm}| p{0.8cm} | p{0.8cm} | p{0.8cm} | p{0.8cm} | p{0.8cm} |} 
 \hline
 \textbf{Metric} & \textbf{Deep Patient (sparse)} & \textbf{Deepr (language)} & \textbf{GRU (language)} & \textbf{GRU (sparse)} & \textbf{BEHRT} & \textbf{LR} \\ [0.5ex]
 \hline
 Sensit - ivity & 0.9172 $\pm$ 0.0183 & 0.9171 $\pm$ 0.0110 & 0.9259 $\pm$ 0.0062 & 0.9396 $\pm$ 0.0041 & \textbf{0.9660 $\pm$ 0.0029} & 0.9435 $\pm$ 0.0003 \\
 \hline
 Specif - icity & 0.6952 $\pm$ 0.04215 & 0.7350 $\pm$ 0.0246 & 0.7376 $\pm$ 0.0125 & \textbf{0.7524 $\pm$ 0.0091 }& 0.6898 $\pm$ 0.0134 & 0.7175 $\pm$ 0.0008 \\
 \hline
 F1 score & 0.8158 $\pm$ 0.0058 & 0.8327 $\pm$ 0.0030 & 0.8403 $\pm$ 0.0014 & \textbf{0.8567 $\pm$ 0.0015} & 0.8464 $\pm$ 0.0015 & 0.8450 $\pm$ 0.0002 \\
 \hline
 AUC ROC & 0.8062 $\pm$ 0.0122 & 0.8260 $\pm$ 0.0071 & 0.8317 $\pm$ 0.0033 & \textbf{0.8460 $\pm$ 0.0028} & 0.8279 $\pm$ 0.0053 & 0.8305 $\pm$ 0.0003 \\
 \hline
\end{tabular}
\newline
\caption{Machine Learning Results (all-cause mortality) \textmd{This table gives the precision, recall, sensitivity and specificity of each of our 5 models and our baseline logistic regression model for the prediction of all-cause mortality. Results are given as \textit{mean $\pm$ standard deviation} with each model being run 10 times using K-fold cross validation. The best result for each metric is given in bold.}}
\label{table:machine_learning_results_all_cause_mortality}
\end{table}

\subsection{Interpretability of Sudden Death}
We include the top 5 features highlighted by each of our 5 models plus our baseline logistic regression model in Table \ref{table:top_5_features_sd}. We can see from this table that models predominantly focus on blood test variables such as alkaline phosphatase and lymphocytes, followed by various prescriptions. Age is identified as important for two of the models built in this study (Deep Patient and our baseline logistic regression model), and only one model identifies a history of a major disease variable as important (BEHRT with history of primary hypertension).

\begin{table}
\centering
\footnotesize
\begin{tabular}{|p{0.1cm} | p{0.9cm} | p{0.9cm} | p{0.9cm} | p{1.0cm} | p{1.0cm} | p{1.0cm} | }  
\hline
\textbf{} & \textbf{GRU (language)} & \textbf{GRU (sparse)} & \textbf{Deep Patient} & \textbf{Deepr} & \textbf{BEHRT} & \textbf{Logistic Regression} \\
 \hline
1 & \color{violet}CRP test & \color{violet}Platelet count test value & \color{violet}eGFR test value & \color{orange}Ant - ibacterial drugs & \color{orange}Drugs used in rheumatic diseases and gout
 & \color{teal}Age \\
\hline
2 & \color{violet}Lympho - cyte count test value & \color{violet}ALT test value & \color{violet}Platelet count test value & \color{violet}ALP test value & \color{orange}Emollient and barrier preparations & \color{orange}General anaesthesia \\
\hline
3 & \color{violet}Mean cell volume test value & \color{violet}AST test value & \color{violet}ALP test value & \color{violet}Lympho - cyte count test value & \color{violet}Nucleated RBC test & \color{violet}ALP test value \\
\hline
4 & \color{violet}Chloride test value & \color{violet}ALP test value & \color{violet}Chloride test value & \color{violet}Diabetes panel in primary care & \color{blue}History of primary hypertension & \color{purple}Encounter for antineoplastic chemo - therapy and immunotherapy \\
\hline
5 & \color{orange}Antidep - ressants & \color{violet}Gamma GT test value & \color{teal}Age & \color{violet}T3 test & \color{violet}HDL test & \color{violet}Red cell distribution width \\
\hline
 \end{tabular}
 \begin{tabular}{|p{0.8cm}|p{1.0cm}|p{1.0cm}|p{1.0cm}|p{1.0cm}|p{1.0cm}|}
 \hline
 \hline
\textbf{KEY}&\color{blue}History of Major Disease & \color{teal}Demo -graphics & \color{purple}Hospital - isation & \color{orange}Presc - ription & \color{violet}Blood test \\
  \hline
 \end{tabular}
 \newline
 \label{table:top_5_features_sd}
\caption{Top 5 Features by Model (sudden death) \textmd{This table gives the top 5 features for each of our 5 models and our baseline logistic regression model, colour coded by type of event. The corresponding colour for each feature family is shown in the key at the bottom of this table.}}
\end{table}

\begin{figure}[!t]
\centering
\includegraphics[width=0.48\textwidth]{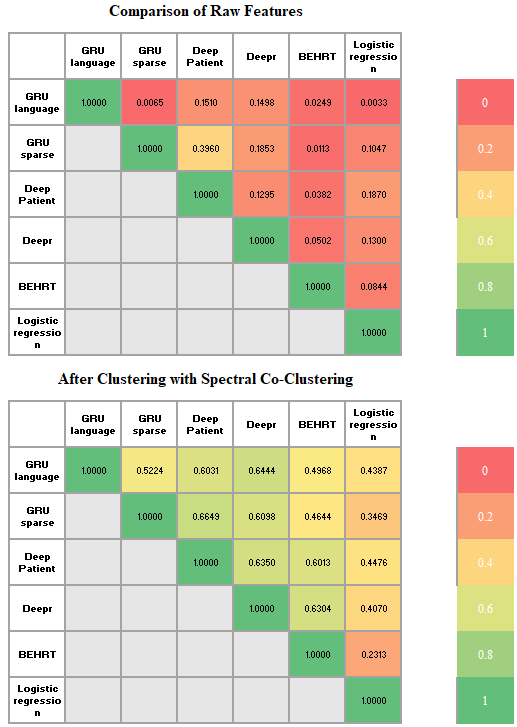}
\caption{Rank Biased Overlap between All-Cause Mortality Prediction Models \textmd{This figure gives a matrix of the RBO between each of our five models plus our baseline logistic regression model. The top figure gives the agreement between raw features, the bottom figure gives the agreement between features after each feature has been replaced by its associated cluster (found using Spectral Co-Clustering). Each table is symmetric along the diagonal, only one side has been given for readability. Cells are colour coded by their value, with red indicating agreement that is closer to 0, and green indicating agreement that is closer to 1.}}
\label{fig:rbo_sd}
\end{figure}

We report the Rank Biased Overlap for the prediction of sudden death and other catastrophic cardiovascular events in Figure \ref{fig:rbo_sd}. We cluster this dataset using Spectral Co-Clustering across the features and patient dimensions into 130 features (chosen using a combination of the Elbow method and attempting to minimise the number of clusters which contain only a single feature). This results in 82 unique feature clusters and 130 unique patient clusters. We report the correlation between each model compared to each other, given as a symmetric 6x6 matrix. When comparing the agreement between raw variables, we can see from this matrix that the highest agreement is between the sparse GRU and Deep Patient (with an agreement of 0.3960), which both use the sparse matrix representation. This is also true when considering correlated features (with an agreement of 0.6649 between the sparse GRU and Deep Patient). We can see from these tables that agreement between models increases significantly when considering the correlation between variables.

\subsection{Interpretability of All-Cause Mortality}
We include the top 5 features highlighted by each of our 5 models plus our baseline logistic regression model for predicting all-cause mortality in Table \ref{table:top_5_features_acm}. As with the sudden death variables, we can see that models consistently identify blood tests as being the most important.

\begin{table}
\centering
\footnotesize
\begin{tabular}{|p{0.1cm} | p{0.9cm} | p{0.9cm} | p{0.9cm} | p{1.0cm} | p{1.0cm} | p{1.0cm} | }  
\hline
\textbf{} & \textbf{GRU (language)} & \textbf{GRU (sparse)} & \textbf{Deep Patient} & \textbf{Deepr} & \textbf{BEHRT} & \textbf{Logistic Regression} \\
\hline
1 & \color{violet}Full blood count panel in primary care & \color{violet}Platelet count test value & \color{violet}Mean cell haemo - globin test value & \color{violet} Other test done in primary care & \color{violet}Calcium test & \color{purple}Lung cancer hospitalisation (unspecified) \\ 
\hline
2 & \color{orange}Wound management prescription & \color{violet}ALT test value & \color{violet}Mean cell volume test value & \color{violet}Other test & \color{violet}Renal blood test panel in primary care & \color{teal}Age \\
\hline
3 & \color{orange}Laxative prescription & \color{violet}Other test & \color{violet}Platelet count test value & \color{orange}Anti-secretory drugs and mucosal protectants prescription & \color{orange}Emollient and barrier preparations prescription & 
\color{purple}Secondary lung cancer hospitalisation \\
\hline
4 & \color{purple}Unspec - ified hospitalisation & \color{violet}CRP test value & \color{violet}ALP test value & \color{orange}Analgesics prescription & \color{violet}Other blood test panel done in primary care & \color{purple}Secondary liver and bile duct cancer hospitalisation \\
\hline
5 & \color{violet}Lymph - ocyte count test value & \color{teal}Age & \color{violet}Haemo - globin test value & \color{violet}eGFR test value & \color{violet}Other blood test & \color{violet}ALP test value \\
\hline
 \end{tabular}
 \begin{tabular}{|p{0.8cm}|p{1.0cm}|p{1.0cm}|p{1.0cm}|p{1.0cm}|p{1.0cm}|}
 \hline
 \hline
\textbf{KEY}&\color{blue}History of Major Disease & \color{teal}Demo -graphics & \color{purple}Hospital - isation & \color{orange}Presc - ription & \color{violet}Blood test \\
  \hline
 \end{tabular}
 \newline
\caption{Top 5 Features by Model (all-cause mortality) \textmd{This table gives the top 5 features for each of our 5 models and our baseline logistic regression model, colour coded by type of event. The corresponding colour for each feature family is shown in the key at the bottom of this table.}}
\label{table:top_5_features_acm}
\end{table}

\begin{figure}[!t]
\centering
\includegraphics[width=0.48\textwidth]{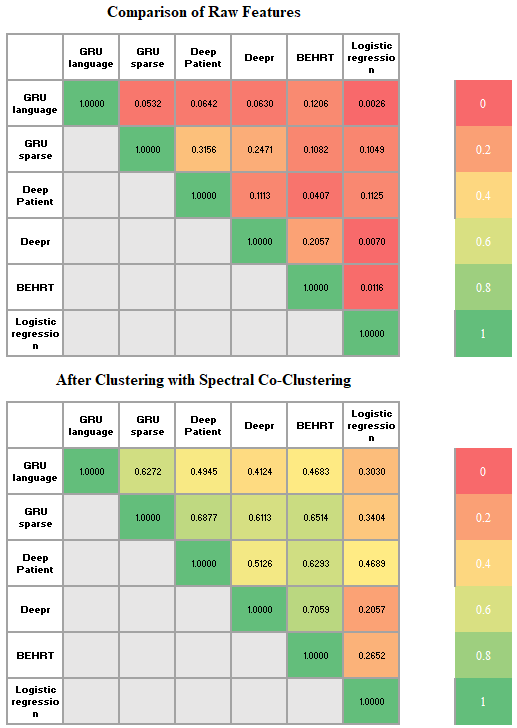}
\caption{Rank Biased Overlap between All-Cause Mortality Prediction Models \textmd{This figure gives a matrix of the RBO between each of our five models plus our baseline logistic regression model. The top figure gives the agreement between raw features, the bottom figure gives the agreement between features after each feature has been replaced by its associated cluster (found using Spectral Co-Clustering). Each table is symmetric along the diagonal, only one side has been given for readability. Cells are colour coded by their value, with red indicating agreement that is closer to 0, and green indicating agreement that is closer to 1.}}
\label{fig:rbo_acm}
\end{figure}

We report the Rank Biased Overlap for the prediction of all-cause mortality in Figure \ref{fig:rbo_acm}. We report the correlation between each model compared to each other, given as a symmetric 6x6 matrix. As with the sudden death results in Figure \ref{fig:rbo_sd}, it appears that the highest agreement between features when considering raw features is with the sparse GRU and Deep Patient (achieving an agreement of 0.3156), both of which use the sparse matrix representation. However, unlike the sudden death results, the highest agreement when considering correlated features is between BEHRT and Deepr (achieving an agreement of 0.7059), both of which use the language representation. For both the sudden death results and the all-cause mortality results, the agreement between models increases considerably when taking correlated variables into account.

\subsection{Comparison of Outcomes}
We now go on to compare the features when predicting sudden death (a composite outcome defined as both sudden death and other catastrophic cardiovascular events) versus all-cause mortality.

\subsubsection{Cumulative Distribution Plot}
As a result of our feature importance analysis, we get a list of features (ordered from most to least important) for each model under test. Each of these features has an associated feature importance score which we normalize across models to find how important each feature is, relative to each other.

With this information, we produce a cumulative distribution plot to determine how many features are needed by each model in order to predict adverse events.

\begin{figure}[!t]
\centering
\includegraphics[width=0.48\textwidth]{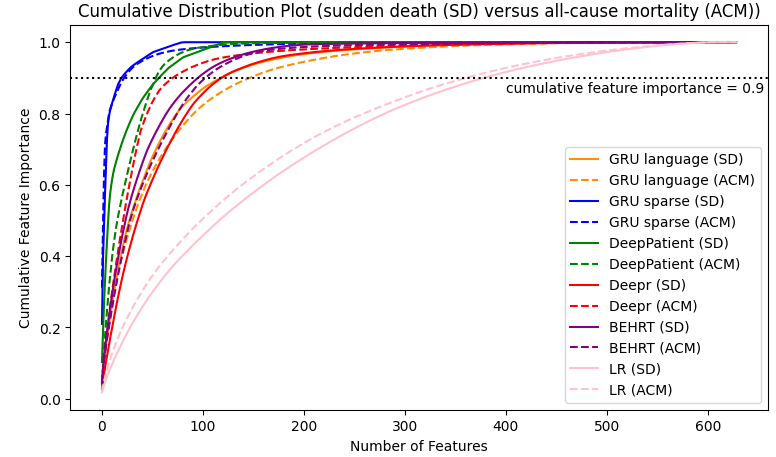}
\caption{Cumulative Distribution Plot: \textmd{This figure gives a cumulative distribution plot of feature importance coefficients for each of our 5 models plus our baseline logistic regression model, for sudden death (solid line) compared to all-cause mortality (dashed line). A horizontal dotted black line shows where the cumulative feature importance reaches 90\% of total feature importance coefficients.}}
\label{fig:cdp}
\end{figure}

We show the cumulative distribution plot for all models under test in Figure \ref{fig:cdp}. This plot gives us an indication of how much each feature contributes to the final feature importance when predicting sudden death and other catastrophic cardiovascular events compared to all-cause mortality. We can see from this plot that for all models, few features contribute highly to the prediction of adverse events, while most have no effect. This is with exception to our baseline logistic regression model, which seems to use all features when predicting both sudden death and all-cause mortality.

\subsubsection{Agreement Between Important Features}

We use spectral co-clustering to group similar features in order to account for the correlated variables in the dataset. We demonstrate the statistical similarity between the top features by calculating the Rank Biased Overlap in Figures \ref{fig:rbo_sd} and \ref{fig:rbo_acm}. We find that agreement between features across models is much higher when considering these correlated variables, as expected. From Figures \ref{fig:rbo_sd} and \ref{fig:rbo_acm} we find the following:

\begin{itemize}
	\item Max agreement between models when predicting sudden death without clustering: 0.360 (Deep Patient and the sparse GRU) 
	\item Max agreement between models when predicting sudden death with clustering: 0.6649 (Deep Patient and sparse GRU) 
	\item Max agreement between models when predicting all-cause mortality without clustering: 0.3156 (Deep Patient and sparse GRU)
	\item Max agreement between models when predicting all-cause mortality with clustering: 0.7059 (BERHT and Deepr)
\end{itemize}

Finally, we compare the agreement between models when predicting sudden death and other catastrophic cardiovascular events to the same models predicting for all-cause mortality, shown in Table \ref{table:acm_vs_sd_interpretability}. We can see from this table that models developed to predict sudden death use different features than models developed to predict all-cause mortality, with some overlap between them. The BEHRT and Deep Patient models identify the most similar features between the two outcomes, achieving an agreement of 0.7390 and 0.7096, respectively.

\begin{table}
\centering
\begin{tabular}{|p{0.8cm} | p{0.8cm} | p{0.8cm} | p{0.8cm} | p{0.8cm} | p{0.8cm} | p{0.8cm} |}
\hline
\textbf{Model} & \textbf{GRU language} & \textbf{GRU sparse} & \textbf{Deep Patient} & \textbf{Deepr} & \textbf{BEHRT} & \textbf{Logistic Regression} \\
\hline
Agree - ment (RBO) & 0.3750 & 0.6304 & 0.7096 & 0.5308 & 0.7390 & 0.4955 \\
\hline
\end{tabular}
\newline
\caption{Sudden Death versus All-Cause Mortality: agreement between models \textmd{This table gives the agreement between feature importance coefficients for the prediction of all-cause mortality versus sudden death for each of the six models we build in this study, using the Clustered Ranked-Biased Overlap method.}}
\label{table:acm_vs_sd_interpretability}
\end{table}

\subsection{Clustering Results}
As discussed, we cluster the dataset used in this study into 130 clusters using Spectral Co-Clustering. We test the stability of this clustering by bootstrapping this dataset 20 times, clustering the resulting dataset, and comparing the clusters produced in each iteration. For each iteration, we use Rank-Biased Overlap to compare the similarity between cluster labels with respect to each feature. We find an average similarity of 0.9522 between resulting cluster groups, indicating the stability of the clustering algorithm.

The spectral co-clustering algorithm used in this study clusters both features and patients simultaneously. We discuss both in the following sections.

\subsubsection{Per-Feature Clustering}
\begin{figure}[t!]
\centering
\includegraphics[width=0.48\textwidth]{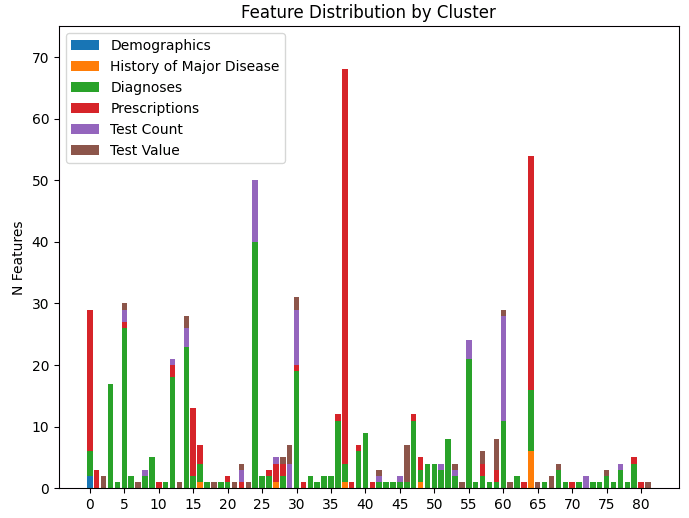}
\caption{Feature Distribution by Cluster \textmd{This plot visualises both the length and contents of each of the feature clusters. Clusters are broken down by feature type, namely demographics (blue), history of major disease (orange), diagnoses (green), prescriptions (red), blood tests (purple), and blood test values (brown). Cluster indices are given along the x-axis.}}
\label{fig:feature_distribution_by_cluster}
\end{figure}

\begin{figure}[t!]
\centering
\includegraphics[width=0.48\textwidth]{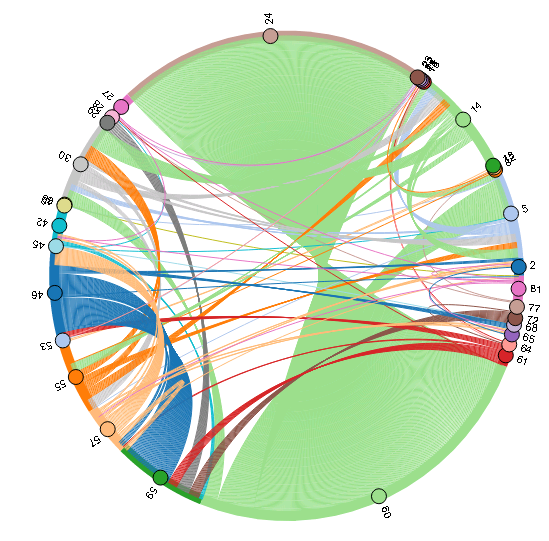}
\caption{Circular Connectivity Plot \textmd{This plot shows connections between features within clusters. The cluster indices are shown around the perimeter of the plot, these are the same indices as the plot in Figure \ref{fig:feature_distribution_by_cluster}. A connection between cluster indicates there exists two features which have a correlation > 0.5 within these clusters. Colours are shown for clarity of reading to separate clusters.}}
\label{fig:circular_connectivity_plot}
\end{figure}

We show the length of each cluster, as well as the feature families contained within each cluster in Figure \ref{fig:feature_distribution_by_cluster}. We show the within-cluster correlations in Figure \ref{fig:circular_connectivity_plot}, where each connection between clusters indicates a pair of features within both clusters that has a correlation of $>$ 0.5. We can see from this diagram that clusters 60 and 24 are strongly connected. Looking into these clusters in more depth, we find that there are 80 pairs of features between these clusters that have a correlation $>$ 0.5 and 13 pairs of features with a correlation $>$ 0.9. We can see from Figure \ref{fig:feature_distribution_by_cluster} that both clusters contain a high proportion of 'Test Count' features within them. It is these features which are causing a strong correlation between clusters, with variables which are typically clustered together (such as mean cell volume and haemoglobin which are always measured together using the same blood test) occurring in both clusters, causing a strong link between them.

\subsubsection{Per-Patient Clustering}
\begin{figure}[t!]
\centering
\includegraphics[width=0.48\textwidth]{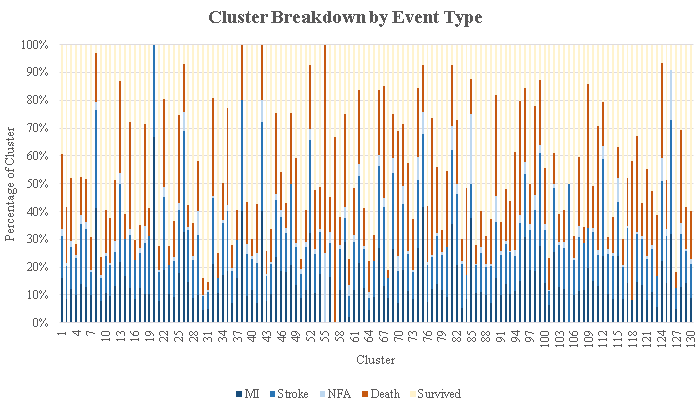}
\caption{Clusters by Event \textmd{This plot shows the distribution of event across each cluster. The x-axis shows each cluster (labelled 1 - 130), the y-axis shows the percentage of events for each cluster.}}
\label{fig:event_by_cluster}
\end{figure}

In our machine learning analysis, we attempt to predict two outcomes within 6 months of index date: no event, and event. Our 'event' outcome contains multiple different types of events: sudden death, and major adverse cardiovascular events (broken down into hospitalisation for myocardial infarction, hospitalisation for stroke, and hospitalisation for near-fatal arrhythmia). We can therefore break down the 'event' group into sub-categories based on the type of event, which we show by cluster in Figure \ref{fig:event_by_cluster}. Most clusters include a combination of all five event types, but all in different proportions. Clusters 30 and 31 contain a far greater proportion of survivors compared to other clusters, at over 85\%. Cluster 30 contains $>$50\% of the total population in this study, and seems to be capturing the majority of those that did not have an event. In comparison, clusters 20, 38, 42, and 55 contain only those that had an event, and clusters 8, 26, 52, 75, 81, 124 and 126 are made up of over 90\% of people with an event. Comparing event type, clusters 20 and 126 contain only people that survived (they either did not have an event, or they survived a hospitalisation for major adverse cardiovascular events).

\section{Discussion}
When building machine learning models for healthcare applications, algorithms need to be compared to appropriate baselines or standards to ensure the performance of these models is high enough to warrant the cost of deploying and maintaining them \cite{RN413}. Performance of models is typically evaluated in relation to other outcomes \cite{RN550,RN377,RN558,RN560,RN561,RN563} or other prediction models \cite{RN550,RN538,RN553,RN548,RN247,RN551,RN549}. To this end, we test the models built in this study which predict sudden death and other catastrophic cardiovascular events in two ways. We recreate three current state-of-the-art models on the Safe Haven platform: BEHRT, Deepr, and Deep Patient. We train each to predict sudden death and other catastrophic cardiovascular events and compare results across various metrics. We additionally train each to predict all-cause mortality, an outcome more commonly predicted by other researchers. We compare two data representations of patient healthcare records: a sparse temporal matrix and a language representation. We find that all-cause mortality is easier to predict, with a mean AUC of 0.83 across models versus 0.75 for the prediction of sudden death.

When defining sudden death, we exclude people that have evidence of a terminal illness, people for whom it is easier to predict for adverse events compared to the general population. This group is likely making the prediction of all-cause mortality easier than the prediction of sudden death since these people are included in our all-cause mortality population. This results both in fewer people in our sudden death group and creating a population in which is harder to assess risk.

Tables \ref{table:top_5_features_sd} and \ref{table:top_5_features_acm} show the top 5 features that are highlighted by each model for the prediction of sudden death and all-cause mortality, respectively. It is clear that the models focus on blood tests and medications disproportionately compared to the other variables. Comparing the features highlighted by sudden death models to those highlighted by models predicting all-cause mortality, we can see that many of the features being used to predict all-cause mortality do not appear in the sudden death dataset, because they are features associated with terminal illness. The hospitalisation variables which are being picked up are generally cancer diagnoses (shown in Table \ref{table:top_5_features_acm}), often a serious problem associated with high mortality. For this reason, we exclude people with terminal illness from our definition of sudden death, making this population harder to predict.

We now go on to compare our results to other papers which have predicted all-cause mortality using EHRs. In 2021, Wang et al \cite{RN562} used EHRs from a hospital in Shanxi Provence, China to predict all-cause mortality within 3 years in patients with heart failure. Authors compared shallow-learning models commonly used by other researchers to an XGBoost classifier. The XGBoost model outperformed all other classifiers, achieving an AUC of 0.82 on their dataset. Authors then used SHAP to interpret their classifier, and found age, NTproBNP, NYHA class, diameter of right atrium, and occupation to be most predictive of all-cause mortality.

In 2022, Qiu et al \cite{RN561} used the National Health and Nutritional Examination Survey (NHANES) dataset to predict all-cause mortality at 1-, 3-, 5-, and 10-years in the general population. The NHANES dataset includes questionnaires, physical examinations, and demographic information of adults aged 20 or older. Using this dataset, authors predicted all-cause mortality using various tree-based machine learning models, achieving an AUC of 0.92 when predicting all-cause mortality within 1 year using a gradient-boosted trees classifier. Similar to Wang et al \cite{RN562}, Qiu et al also used SHAP to interpret their models, reporting age, red cell distribution width, arm circumference, urine albumin, and "general health" as most predictive of all-cause mortality according to these techniques. However, it is likely the NHANES dataset used by Qiu et al suffers from the healthy volunteer effect; volunteers for medical research tend to be healthier than the average population \cite{RN326,RN327}.

In this study, we use records from the NHS GG\&C Safe Haven dataset to predict both all-cause mortality and sudden death within 6 months in people aged 50 or older. Unlike the work by Wang et al \cite{RN562}, we predict for these events across the entire population, rather than those that are already suffering from major illness. Since this dataset includes records collected throughout NHS services in everyone living within the GG\&C area, it should be more representative of the general population than the NHANES dataset used by Qiu et al \cite{RN561}.

Unlike Wang et al and Qui et al who both use SHAP to interpret their models, the authors of BEHRT \cite{RN261} use local explainability from attention coefficients to interpret their model. BEHRT is a self-attention based transformer model, and so by visualising the attention weights we can capture some aspects of local explanations. This is the same approach used by many researchers for visualising the original BERT language model \cite{RN424}. However, there has recently been some criticisms of this approach. Fenech, M. and Buston, O. \cite{RN421} analyse the attention weights versus model behaviour for attention-based architectures across a variety of tasks, focusing on faithfulness of explanations and stability of results. The authors conclude that attention is not a reliable measure of feature importance, finding that the same prediction can be produced with wildly different attention distributions, breaking the stability property.

Both global and local interpretability are important for explaining model decisions, it is local interpretability that can inform a clinician about the reason for a model prediction for an individual patient, which might provide key information enabling the clinician to intervent to prevent the event. However, the risk of misunderstanding the interpretation to both end-users and data scientists has been found to be heavily associated with local interpretability methods \cite{RN194,RN203,RN210}. These methods only give a tiny snapshot on model thinking, leading to incorrect assumptions about the data and a false sense of security \cite{RN194}.

In this study, we focus on global interpretability (as opposed to local interpretability) to understand model thinking and extract insights of model behaviour. However, in some of the models built in this study, we calculate local interpretations and average these across the dataset to extract global insights from them, which is the same approach done by Rao et al in 2022 \cite{RN522} and Jones et al in 2020 \cite{Jones_2020}. However, due to the prohibitively high computational complexity and size of the dataset used in this study, we chose not to apply interpretability techniques such as SHapley Additive exPlanations (SHAP) \cite{NIPS2017_8a20a862} as used by Qui et al and Wang et al. Instead, we focused on coupling local-interpretability with co-clustering techniques to provide insight about the consensus across prediction models and tasks. Due to the highly sensitive nature of this healthcare dataset, we are unable to provide details of local interpretations mentioned throughout this study as this would disclose sensitive patient information.

We now go on to further interpret the models built in this study by grouping features into feature families and determining which families contribute most to model predictions through an Ablation Study (see Appendix J). From this analysis, we find that the highest model AUC is found when all 629 features are included in this study. A similar AUC is found when just diagnoses, demographics and prescriptions are included in the model. This is the same finding as reported by both Roberto, J. et al \cite{RN262} and Rao, S. et al \cite{RN522}. Even though datasets and outcome being predicted differ between these studies and ours, this outcome shows the usefulness of simple observational data such as prescriptions and hospital records for the prediction of a variety of outcomes.

Finally, we go beyond the work by other researchers and attempt to compare how much these state-of-the-art models agree with each other when ranking important features for the prediction of either all-cause mortality or sudden death. To account for correlated variables in the dataset used in this study, we use Spectral Co-Clustering to group patients and features simultaneously, allowing us to group features with regards to specific patient sub-populations. We find grouping the dataset used to predict sudden death into 130 features yields the best results for this purpose, as we discuss in more depth in Appendix K. As with the raw Rank-Biased Overlap results, we also find agreement between models to be higher when predicting for sudden death compared to all-cause mortality when considering clustered variables importance coefficients.

\subsection{Limitations}
There are some limitations associated with this study. Firstly, the dataset used in this paper includes only people aged 50 or older as of the 1st of January 2010, meaning we are unable to test our prediction models on those aged younger than 50 in whom rates of events are typically much less common and even harder to predict. Similarly, the dataset used in this study only contains people living in the GG\&C region. The features identified as important in this dataset may differ when considering other cities or countries due to the high levels of social deprivation in this population \cite{RN457,RN483,RN458}.

Additionally, we are unable to get access to the datasets used by some of the other state-of-the-art studies we reference in this research, namely CPRD used in BEHRT \cite{RN261}. CPRD contains a different set of features and a different level of granularity compared to the Safe Haven dataset, and therefore we are unable to test the exact models proposed by the original researchers, and instead rely on being able to recreate their model for use in this dataset.

\section{Conclusion}

In this paper, we have built six machine-learning models adapted from state-of-the-art research for the prediction of sudden death using EHRs. As far as we are aware, this is the first paper to predict sudden death on such a large dataset (population ~300,000) in apparently healthy people.

In doing so, we achieve an AUC of around 0.75 compared to an AUC of around 0.83 for all-cause mortality.

Following the work of other researchers, we use interpretability techniques to compare important features when predicting both sudden death and all-cause mortality. Unlike other work in this field, we use Rank-Biased Overlap to compare how much each model agrees with the other when considering important features across both outcomes. As far as we are aware, we are the first to examine the consensus of interpretability results across models and tasks for the prediction of events within electronic health records.

We find that when looking at the raw features used by the models, the agreement between models is poor. Spectral Co-Clustering to group features and patients simultaneously gave a better indication of the relationship between features than correlation alone.

This research emphasises the lack of useful risk prediction models for sudden death in the general population, and outlines the challenges when applying machine learning models to sparse, retrospective datasets like electronic health records. We emphasise the need for rigorous comparisons between models, both when comparing raw model performance, and when considering the interpretation of model behaviour.


%

\appendices
\section*{Appendix A: Outcome Definition}
In this study, we predict for both all-cause mortality and sudden death within 6 months of the index date, given the past year of patient medical records. For our outcome groups, this index date becomes a random date up to 6 months (between 0 and 180 days) before date of death. Our control group when predicting sudden death and other catastrophic cardiovascular events is a group of people in the SafeHaven dataset who are not already sick; this index date is a randomly selected date in which the individual did not suffer a catastrophic cardiovascular event within the following six months, although they could have such an event after a period of $>$ 6 months.

\begin{figure}[t!]
\centering
\includegraphics[width=0.48\textwidth]{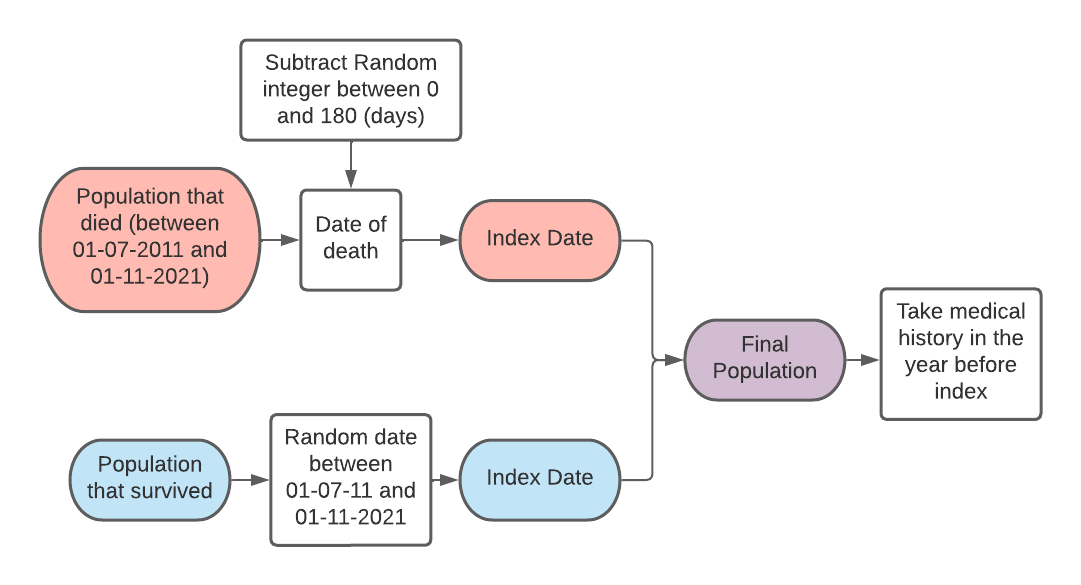}
\caption{Population Definition Flow Diagram. This figure gives a flow diagram of how we convert a control and event group to a training and testing set which can be used by our machine learning models.}
\label{fig:pop_def}
\end{figure}

A flow diagram of how we define our sudden death and other catastrophic cardiovascular event population can be found in Figure \ref{fig:pop_def}.

\section*{Appendix B: Data Cleaning}

Since each table in this dataset uses a different coding system and comes from a different source, the sources of noise in each differ. We therefore describe how we clean each of our tables in the following sections.

\subsection*{Lab Tests}
The blood test dataset is the noisiest of the datasets we work with, and therefore requires the most cleaning. There are two important features to consider

\begin{itemize}
    \item The fact that the test was done. This may be done because the person has developed symptoms and sign requiring investigation and one abnormal test may lead to the need for additional tests. Alternatively, tests may be done to screen for disease in otherwise healthy people or done routinely prior to a procedure or as part of the follow-up of known disease. Unfortunately, the reasons why a test is done is not recorded, although the circumstances of the test may be helpful (eg:- during hospitalisation, associated with hospital clinic visit etc).
    \item The results of the test. A blood test results may be normal or abnormal. The severity of the abnormality may predict outcome.
\end{itemize}

We describe the sources of noise below.

\begin{itemize}
    \item The variable under test is given as free-text, and therefore sources of noise include typos, abbreviations, synonyms, and inconsistent capitalisation
    \item The clinical code value used to sort each test is also prone to sources of noise. These include 1s being entered as capital Is or lowercase ls (or vice versa), 0s being represented as a capital or lowercase O letter O, and the value being given as both abbreviations or chemical compositions where relevant (e.g. bicarbonate is given as both the abbreviation 'BIC' and the chemical composition 'HC03'
    \item As with the variable under test, the units under test are given as free-text, and therefore are affected by the same sources of noise
    \item The numerical value of the test is also prone to mis-entry. Common sources of noise include missing data being entered as zeros (especially disruptive when the result of the test could actually be zero, for example in the case of thyroid-stimulaing hormone, TSH), missing or misplaced decimal points, or when the unit for the test does not match the given value
\end{itemize}

The SCI Store dataset contains just under 100 million entries, cleaning all of these would be infeasible. Therefore, we limit our analysis to a certain group of blood test variables. We choose these variables based on the following:

\begin{enumerate}
    \item By finding the most common blood tests taken
    \item Through clinician expertise, by asking clinicians what tests they would look out for when making a call on whether a person was at risk for major adverse events
\end{enumerate}

Any test not included in the list of tests we're interested in is grouped into a final 'other' category. 

To clean the dataset, we do the following:

\begin{algorithm}
\begin{algorithmic}
  \State We filter out only the blood tests using the tissue type column
  \For{each variable we're interested in}
    \State Create a lookup table of all names for that test which appear in the dataset (experimenting with typos, synonyms, abbreviations)
    \State Find all unique units of measurement, and create conversion conditions for each to make sure each test is in the same units (for example $g/L * 1000$ becomes $mg/L$)
    \State Create a 'maximum biologically possible' and 'minimum biologically possible' value (found by asking clinical experts)
    \State Label every value that appears over this biological maximum or under the biological minimum as 'missing'
  \EndFor
\end{algorithmic}
\end{algorithm}

This does not filter out all possible sources of noise, for example missing data being encoded as zero for variables in which zero is biologically possible, however does give a cleaner subset of the data.

\subsection*{Hospitalisations}
Hospital diagnoses in this dataset are given as ICD-10 codes and are assigned to a record on discharge. In the case of ICD-10 codes, each code is represented by a letter and followed by roughly 4 digits. This leads to roughly 86,000 unique ICD-10 codes. If we were to use the complete ICD-10 code to represent each of the six possible diagnosis codes, we'd need a feature vector of approximately size 516,000 in order to represent six ICD-10 codes using one-hot encoding.

We therefore want to reduce the dimensionality of the data. We do the same thing as Meng et. al do in \cite{RN289}, and take the first 3 digits of the ICD-10 code. This reduces the size of our feature embedding from 86,000 per ICD-10 code, to roughly 80 per ICD-10 code. For example, ICD-10 code I26.02 (saddle embolus of pulmonary artery with acute cor pulmonale) becomes ICD-10 code I26.0 (pulmonary embolism with acute cor pulmonale).

We define ’length of hospital stay’ as  the number of days between  discharge date and admission date. For appointments   in which a patient arrives and is discharged on the same day, length of stay was 0 days.  For hospital transfers, we define admission date as the admission date of the first hospital, and discharge date as the discharge date from the last hospital. We define diagnoses of these visits as the diagnosis codes assigned to the last hospital stay after the patient has been transferred.

\subsection*{Pharmacy}
Prescriptions in this dataset are given as BNF codes, where each medication is represented by an alphanumeric code of up to 10 digits long. To reduce the number of features required to store each unique prescription, we cut the BNF code at between 4 and 6 digits (giving the medication chapter, section, and paragraph) to reduce the dimensionality of our data.

\section*{Appendix C: Text Vectorization Methods}
Two possible text vectorization methods are integer encoding and bag-of-words encoding. Both vectorization methods require a 'vocabulary', a set of unique words which could appear in the input text. The vocabulary used depends on the application; for a model translating French into English, this will be a set of French and English words. For healthcare applications, this can be ICD-10 codes and medications. Each vocabulary usually also contains an "unknown" token (which we refer to as [UNK], to account for words that are given to the model but do not appear in the vocabulary).

The bag-of-words encoding turns the language representation into a sparse vector, with each ‘feature’ in this representation being an event of interest. In contrast, in integer encoding each ‘feature’ in this representation is a sentence position. This is important when we consider interpretability. When considering global interpretability, we attempt to understand what features matter in the model. We do this by manipulating the values at each feature in the training or testing set. In integer encoding, this will give us the position in the sentence that matters most. For bag-of-words encoding, this gives us the words in the sentence that matter most, which we typically care about more than positional importance. Therefore, an extra step is required for integer encoding in order to generate event importance indices.

In integer encoding, each word in the vocabulary is assigned to a unique integer. We replace each word in the input sentence with its corresponding integer representation, creating a vector of integers for each sentence. In contrast, for bag-of-words representation, sentences are represented as how many times each word in the vocabulary appears in the sentence.

\section*{Appendix D: Feature Processing}
In the following section, we describe how we create both our language model and sparse matrix data representations from individual feature tables within the NHS Safe Haven dataset.

\subsection*{Feature Processing}
\begin{itemize}
    \item \textbf{Prescriptions} In the NHS Greater Glasgow and Clyde Safe Haven dataset, prescriptions are represented as BNF codes; we represent each medication as the first four digits of the BNF code.
    \item \textbf{Hospitalisations} In the NHS Greater Glasgow and Clyde Safe Haven dataset, hospital records contain up to six diagnoses for each visit, with each diagnosis given as an ICD-10 code. We take the first two diagnoses (primary and secondary diagnosis for each visit) and represent each diagnosis as the first four digits of the ICD-10 code.
    \item \textbf{Blood tests} The blood test dataset is the noisiest dataset we work with in this study, and therefore requires the most pre-processing. We group blood tests into 42 unique categories (41 blood test types, e.g. haemoglobin, calcium, etc, plus 'other'). We capture six types of information from the blood test dataset:
    \begin{itemize}
        \item A marker that a test was taken (e.g. calcium was tested)
        \item The result of each test (numerical, e.g. 2.2 mmol/L)
        \item How many samples of blood were taken in hospital for each patient (secondary care)
        \item How many samples of blood were taken by a GP for each patient (primary care)
        \item We then group our 41 variables into eight blood test families: full blood count, diabetes, renal, bone and liver, thyroid, cholesterol, inflammation, and coagulation. We record how many tests of each group were taken in hospital (primary care) and in hospital (secondary care)
    \end{itemize}
    For the results of each test, we filter out clinically impossible blood tests (e.g. a red blood cell count of zero), and label these values as 'missing'. For the rest of the categories, we include a simple count of how many of each test was taken.
    \item \textbf{Demographics} We record the age and sex at index date for each patient
    \item \textbf{History of major disease} For each patient, we focus on 11 of the following major diseases, and define this as being hospitalised for each disease at any point in the person's history:
    \begin{itemize}
        \item Heart failure
        \item Chronic ischemic heart disease
        \item Liver disease
        \item Seizures
        \item Stroke
        \item Kidney disease
        \item Primary hypertension
        \item Hypertensive heart disease
        \item Chronic kidney disease
        \item Type 1 diabetes
        \item Type 2 diabetes
    \end{itemize}
\end{itemize}

\subsection*{Language Model}

\begin{figure}[t!]
\centering
\includegraphics[width=0.5\textwidth]{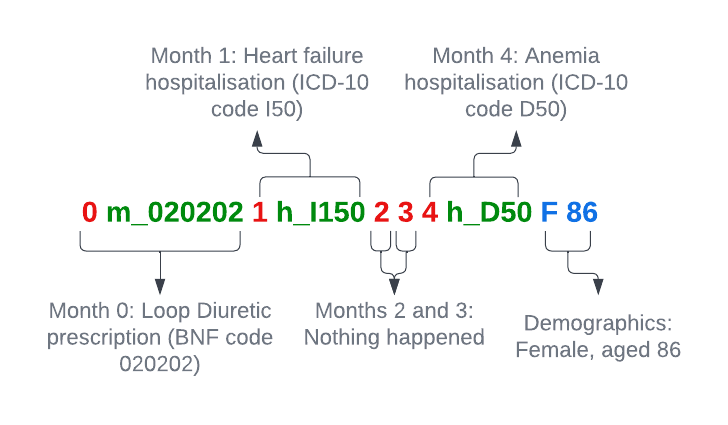}
\caption{Encoded Sentence Example. This figure gives an example sentence which is used to represent patient records. The sentence is split into 'segments' (month 0, 1, 2, 3 and 4). Colours have been used to highlight different types of features; red indicates time, green indicates events, and blue indicates 'static' variables such as age and sex.}
\label{fig:sentence}
\end{figure}

In language models, healthcare records are encoded as sentences which are broken down in to keywords allowing key medical concepts and time-dependent information to be represented. For each hospital diagnosis, medication prescription, and blood test, each code is pre-pended with a feature type identifier ($h\_$, $m\_$ and $t\_$ respectively). Each feature is binned into 2 month intervals, and events are sorted from most to least recent month (from index). Each new month is marked by a $segment\_x$ feature, where x is the number of months working backwards from index. For example, [$segment\_0$ $h\_I50$ $segment\_1$ $m\_0202$] tells us a hospital diagnosis of code I50 (heart failure) was given at the most recent month, with a prescription of code 0202 (diuretics) was given the month previously. We add 2 additional features to the end of each sentence, a feature giving the person's sex, and a feature giving the person's age at index. We then add any history of major disease at the end of each sentence, with each condition prepended by a $hist\_$ identifier. An example of this can be seen in Figure \ref{fig:sentence}.

\subsection*{Sparse Temporal Matrix}
We encode all hospitalisations, medications and blood tests into their distinct categories as discussed previously. Each event is then binned into 2 month chunks, totalling 7 possible chunks between $index$ and $index - 1   year$. We then create a 2D matrix for each patient, of size $n\_features \times n\_time\_windows$. Each patients history is then added to the appropriate cell, where a value of 1 at feature $f$ and time $t$ indicates that an event of type f occurred t months before index. Finally, we add 13 constant features: two demographic columns (age and sex) and 11 variables indicating history of major disease, where each of these features are duplicated for all discrete time intervals. The result is a sparse 2D matrix for each patient which gives a marker of when different types of events occurred in the year before index, plus age and sex. The final array is a 3D matrix of size $n\_patients \times n\_features \times n\_time\_windows$.  The array for all patients is stored in $(i,j,k,v)$ form, where $i$ is the patient, $j$ is the feature, $k$ is the time, and $v$ is the value at that cell. The size of the final matrix is also saved, meaning the dataset can be rebuilt before being passed into each model.

\section*{Appendix E: Model Construction Details}
In this section, we describe implementation details of each of the five models under test in this study.

\begin{itemize}
    \item \textbf{\textit{BEHRT}} As discussed, we train BEHRT by first passing it the entire patient dataset with unsupervised learning, trained using a masked language model. The model is then fine tuned for the prediction of sudden death and other catastrophic cardiovascular events. As with the original BEHRT model, the first layer is a TextVectorizer with integer encoding, fed into a feedforward multi-headed self-attention network, with four layers consisting of 20 heads in each, and an embedding layer of size 256.
    \item \textbf{\textit{Deepr}} Like BEHRT, the first layer of this model is a TextVectorizer with integer encoding. This is fed into a 1D convolutional layer with 45 filters, which is then passed into a max pooling layer, then finally a dense layer for the prediction of the outcome of interest. Like in \cite{RN262} we add a second model input to give the model demographic information (specifically age and sex), via an extra dense layer.
    \item \textbf{\textit{DeepPatient}} For DeepPatient, we use a 3-layer stack of denoising autoencoders for model pre-training using unsupervised learning. The model is trained to reconstruct the original training data which has been corrupted by randomly setting a proportion of the input data to zero (in this case we use 5\%). For the denoising autoencoder, we use 3x 500 neuron fully connected layers as an encoder, followed by a stack of 3x 500 neuron fully connected layers as a decoder. Once the autoencoder has been trained, the decoder is chopped off and the output of the encoder is fed to a downstream classification model (in this case, a GRU).
    \item \textbf{\textit{GRU (language)}} For the language GRU, we use the TextVectorizer as with both BEHRT and Deepr models. Unlike these models, we use a bag-of-words count based encoding to create a sparse vector of size \textit{vocab\_size} for each patient. Our GRU consists of 5 layers; a 100-unit layer and a 50-unit layer with a dropout layer of 10\% and 30\% after each layer respectively. The final dropout layer is fed into a fully connected layer with 2 input neurons.
    \item \textbf{\textit{GRU (sparse)}} The sparse representation of both our GRU is very similar to our language model representation, with the same architecture for both. However, because our sparse matrix is already in machine readable form. we do not need to pass it through a vectorize layer.
\end{itemize}

\section*{Appendix F: Interpretability Techniques}
In the following section we describe the two interpretability techniques used in this paper.

\subsection*{Permutation Feature Importance}
Permutation Feature Importance (PFI) is a method of global interpretability. It measures the difference in model performance after features are perturbed (Algorithm \ref{algorithm:pfi}) \cite{RN278}. If a feature is permuted and the model performance does not change, the feature was clearly unimportant to model decision making. Conversely, if a feature has a large impact on model performance, that feature contributed heavily to model decision making, and therefore the feature is important to the model.

\begin{algorithm}
    \begin{algorithmic}[1]
        \State Calculate original model error $e_{orig}$ using dataset $X$
        \For{each feature $j \in \{1,..,p\}$}
            \State Permute the feature $j$ in dataset $X$ to create a new dataset $X_{perm}$
            \State Calculate model error $e_{perm}$ based on new dataset $X_{perm}$
            \State Calculate feature importance $FI_j = e_{orig} - e_{perm}$
        \EndFor
        \State Sort feature importances $FI$ from most to least important
    \end{algorithmic}
\caption{Permutation Feature Importance Algorithm}
\label{algorithm:pfi}
\end{algorithm}

PFI is a very simple but very powerful tool, as it is easy to implement, easy to rationalise, model agnostic, and provides global interpretability. However, since each feature is permuted individually, it can produce unrealistic scenarios if features are highly correlated, which can affect the validity of global explanations.

\subsection*{LIME}
Local interpretable model-agnostic explanations (LIME) is a technique for generating local interpretations which explain what features a model is interested in when looking at a particular patient \cite{RN260}. It does this by perturbing each of the features in the patient record, passing the modified data back through the original model, then training an intrinsically interpretable model to mimic the behaviour of the more complex model, and interpreting the simpler model. The explanations produced by LIME can be described by the following equation:

\begin{equation}
    \upxi(x) = \underset{g \in G}{argmin}  \mathcal{L}(f,g,\pi_x) + \Omega(g)
\end{equation}

where:
\begin{itemize}
    \item $G$ is a class of interpretable models (e.g. a linear model or decision tree)
    \item $g$ is an intrinsically interpretable model from class $G$
    \item $\Omega(g)$ is a measure of the complexity of the explanation $g \in G$
    \item $f$ is the original "complex" model
    \item $\pi_x$ is a locality measure, how large a neighborhood we consider around datapoint $x$
    \item $\mathcal{L}$ is a fidelity function (how faithful the explanation is to the underlying model)
\end{itemize}

as defined in the original paper \cite{RN260}.

\section*{Appendix G: Statistical Comparison Techniques}
\subsection*{Ranked Biased Overlap}
Rank Biased Overlap (RBO) is an alternative to Kendall Tau correlation for calculating the similarity between two ranked lists. Unlike Kendall-Tau, it weights the head of the list as more important than the tail (the weighting of which is defined by tuneable parameter P). This is especially important in our case, as many of the features in our lists have the same feature importance (zero), and are therefore ordered randomly. RBO is defined as the following equation \cite{RN275}:\\

$$
\mathrm{RBO}(\mathrm{S}, \mathrm{T}, \mathrm{p}, \mathrm{k})=\frac{\mathrm{X}_{\mathrm{k}}}{\mathrm{k}} \cdot \mathrm{p}^{\mathrm{k}}+\frac{1-\mathrm{p}}{\mathrm{p}} \sum_{\mathrm{d}=1}^{\mathrm{k}} \frac{\mathrm{X}_{\mathrm{d}}}{\mathrm{d}} \cdot \mathrm{p}^{\mathrm{d}}
$$\\

where:\\
\begin{itemize}
    \item S and T are two ranked lists under test
    \item d is the depth being examined
    \item k is the max depth of the list
    \item $X_a$ is the size of the overlap between S adn T up to depth a
    \item p is a tuneable parameter which tells the function how much weight to put at the head of the list compared to the tail
\end{itemize}

\section*{Appendix H: Final Feature List}

\begin{itemize}
    \item 318 hospitalisation variables
    \item 196 medication variables
    \item 103 blood test variables
    \item 10 variables indicating history of major disease
    \item 2 demographic variables (age and sex)
\end{itemize}

\section*{Appendix I: Training Curves}
\begin{figure}[t!]
\centering
\includegraphics[scale=0.5]{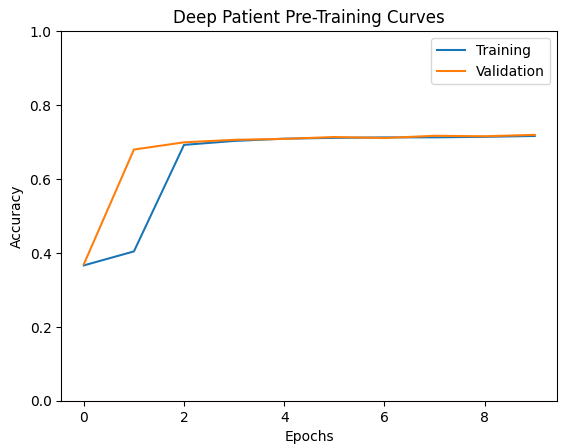}
\caption{DeepPatient Pre-Training Plot. This figure gives the training curves for the pre-training step of the DeepPatient model. Training accuracy is given in blue, validation accuracy is given in orange.}
\label{fig:deeppatient_pretraining}
\end{figure}

\begin{figure}[t!]
\centering
\includegraphics[scale=0.5]{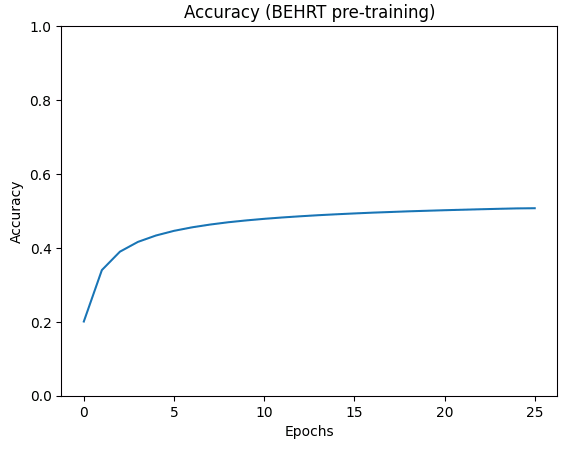}
\caption{BEHRT Pre-Training Plot. This figure gives the training curves for the pre-training step of the BEHRT model.}
\label{fig:behrt_pretraining}
\end{figure}

\begin{figure}[t!]
\centering
\includegraphics[width=0.45\textwidth]{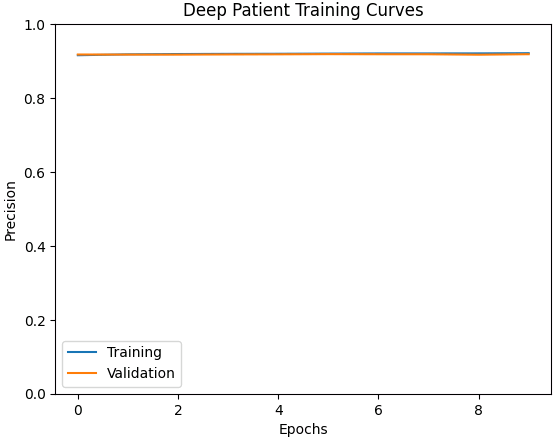}
\includegraphics[width=0.45\textwidth]{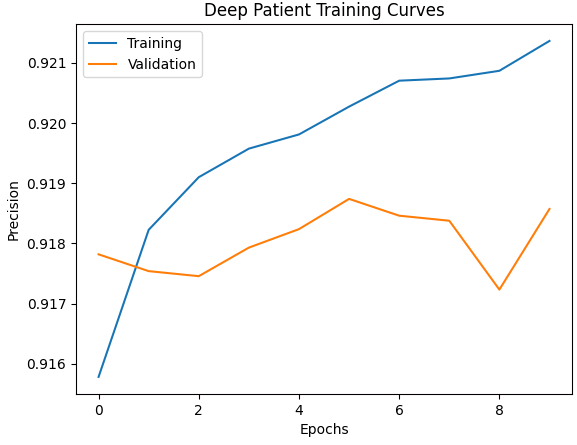}
\caption{Deep Patient Training Curves. This graph gives the training and validation curves for the Deep Patient model trained for 10 epochs. The first plot gives the scale between 0.0 and 1.0, the second plot gives the scale auto-adjusted so we can see how the model trains over time}
\label{fig:deeppatient_training}
\end{figure}

\begin{figure}[t!]
\centering
\includegraphics[width=0.45\textwidth]{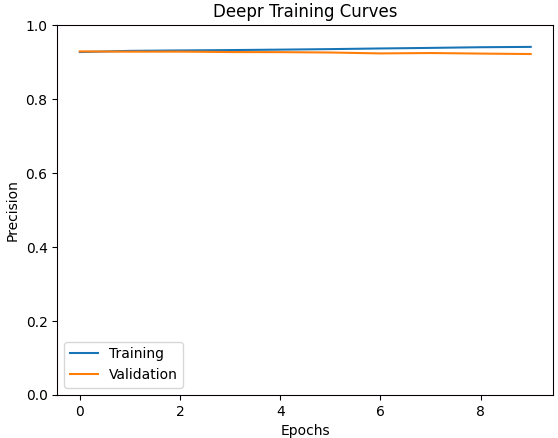}
\includegraphics[width=0.45\textwidth]{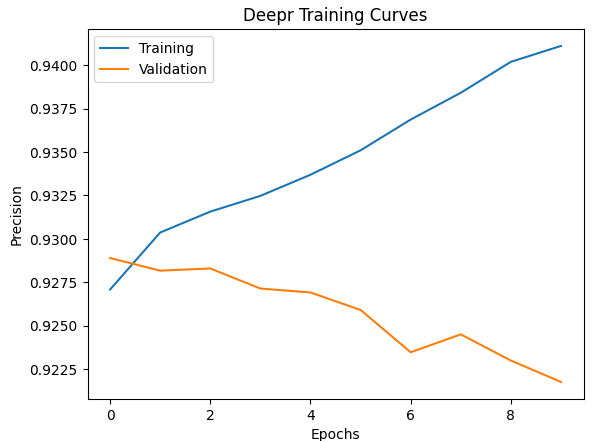}
\caption{Deepr Training Curves. This graph gives the training and validation curves for the Deepr model trained for 10 epochs. The first plot gives the scale between 0.0 and 1.0, the second plot gives the scale auto-adjusted so we can see how the model trains over time.}
\label{fig:deepr_training}
\end{figure}

\begin{figure}[t!]
\centering
\includegraphics[width=0.45\textwidth]{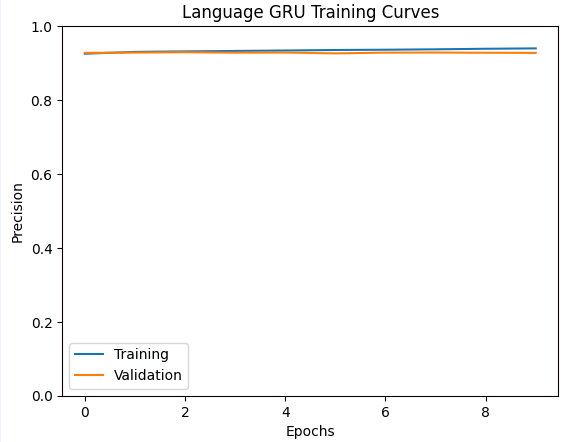}
\includegraphics[width=0.45\textwidth]{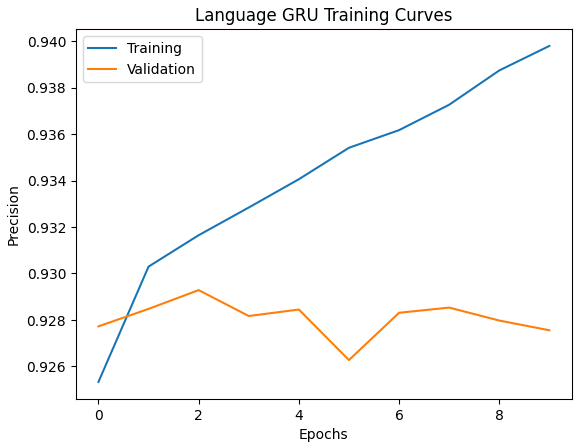}
\caption{Language GRU Training Curves. This graph gives the training and validation curves for the language GRU model trained for 10 epochs. The first plot gives the scale between 0.0 and 1.0, the second plot gives the scale auto-adjusted so we can see how the model trains over time}
\label{fig:language_gru_training}
\end{figure}

\begin{figure}[t!]
\centering
\includegraphics[width=0.45\textwidth]{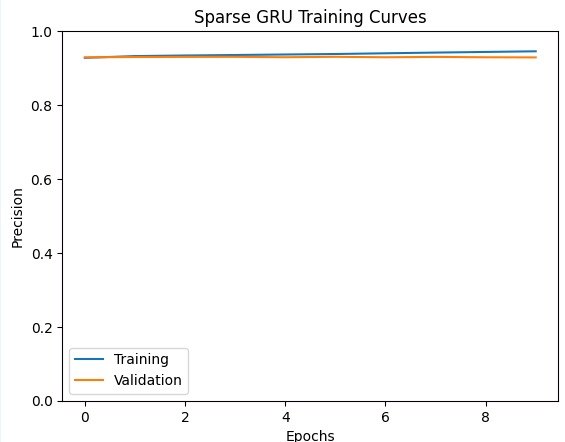}
\includegraphics[width=0.45\textwidth]{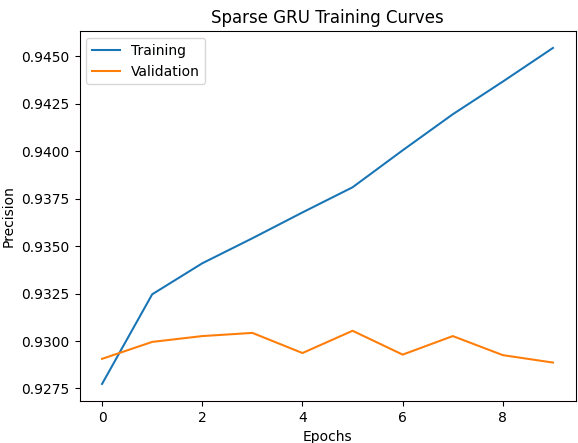}
\caption{Sparse GRU Training Curves. This graph gives the training and validation curves for the sparse GRU model trained for 10 epochs. The first plot gives the scale between 0.0 and 1.0, the second plot gives the scale auto-adjusted so we can see how the model trains over time}
\label{fig:sparse_gru_training}
\end{figure}

\begin{figure}[t!]
\centering
\includegraphics[width=0.45\textwidth]{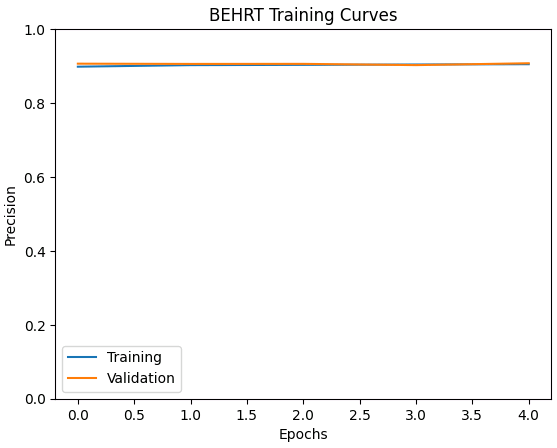}
\includegraphics[width=0.45\textwidth]{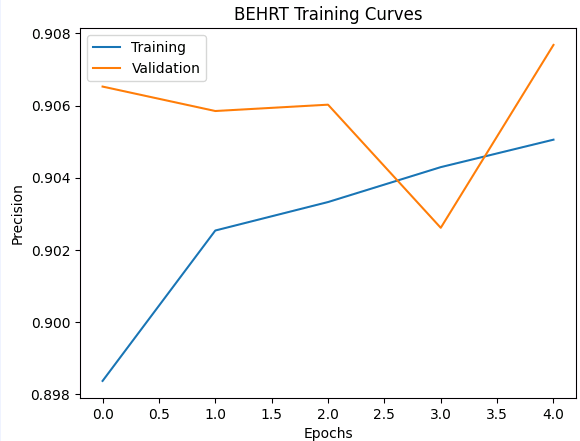}
\caption{BEHRT Training Curves. This graph gives the training and validation curves for the BEHRT model trained for 10 epochs. The first plot gives the scale between 0.0 and 1.0, the second plot gives the scale auto-adjusted so we can see how the model trains over time}
\label{fig:behrt_training}
\end{figure}

\section*{Appendix J: Ablation Study}
We report the results of our ablation study in Table \ref{table:ablation_study}.

\begin{table}
\centering
\begin{tabular}{|p{1.5cm} | p{1.2cm} | p{1.2cm} | p{1.2cm} | p{1.2cm} |} 
 \hline
 \textbf{Feature Subset} & \textbf{Sensitivity} & \textbf{Specificity} & \textbf{AUC} & \textbf{F1 score} \\ [0.5ex]
 \hline
 Demographics & 0.9148 $\pm$ 0.0108 & 0.4311 $\pm$ 0.0270 & 0.6730 $\pm$ 0.0081 & 0.6885 $\pm$ 0.0072 \\
 \hline
 Diagnoses & 0.9639 $\pm$ 0.0054 & 0.2333 $\pm$ 0.0120 & 0.5986 $\pm$ 0.0034 & 0.5988 $\pm$ 0.0055 \\
 \hline
 Prescriptions & 0.9241 $\pm$ 0.0084 & 0.4797 $\pm$ 0.0203 & 0.7010 $\pm$ 0.0062 & 0.7193 $\pm$ 0.0052 \\
 \hline
 History of major disease & 0.9253 $\pm$ 0.0019 & 0.3049 $\pm$ 0.0048 & 0.6134 $\pm$ 0.0015 & 0.6211 $\pm$ 0.0018 \\
 \hline
 Blood test indicator & 0.9528 $\pm$ 0.0091 & 0.2668 $\pm$ 0.0241 & 0.6098 $\pm$ 0.0075 & 0.6149 $\pm$ 0.0111 \\
 \hline
 Blood test value & 0.9437 $\pm$ 0.0157 & 0.3660 $\pm$ 0.0421 & 0.6572 $\pm$ 0.0133 & 0.6735 $\pm$ 0.0147 \\
 \hline
 Demographics + diagnoses & 0.9235 $\pm$ 0.0147 & 0.4926 $\pm$ 0.0349 & 0.7080 $\pm$ 0.0103 & 0.7261 $\pm$ 0.0079 \\
 \hline
 Demographics + diagnoses + history of major disease & 0.9466 $\pm$ 0.0074 & 0.3212 $\pm$ 0.0202 & 0.6339 $\pm$ 0.0064 & 0.6461 $\pm$ 0.0079 \\
 \hline
 Demographics + diagnoses + history of major disease + prescriptions & 0.9326 $\pm$ 0.0088 & 0.4982 $\pm$ 0.0221 & 0.7154 $\pm$ 0.0067 & 0.7354 $\pm$ 0.0052 \\
 \hline
 All features & 0.9253 $\pm$ 0.0083 & 0.6077 $\pm$ 0.0214 & 0.8279 $\pm$ 0.0053 & 0.8464 $\pm$ 0.0015 \\
 \hline
\end{tabular}
\newline
\caption{Ablation Study Results \textmd{This table gives the results of a sparse GRU trained on a subset of features included in this study. The subset of features under test is given in the lefthand column.}}
\label{table:ablation_study}
\end{table}

\section*{Appendix J: Optimum Number of Clusters using the Elbow Method}
\begin{figure}[t!]
\centering
\includegraphics[width=0.48\textwidth]{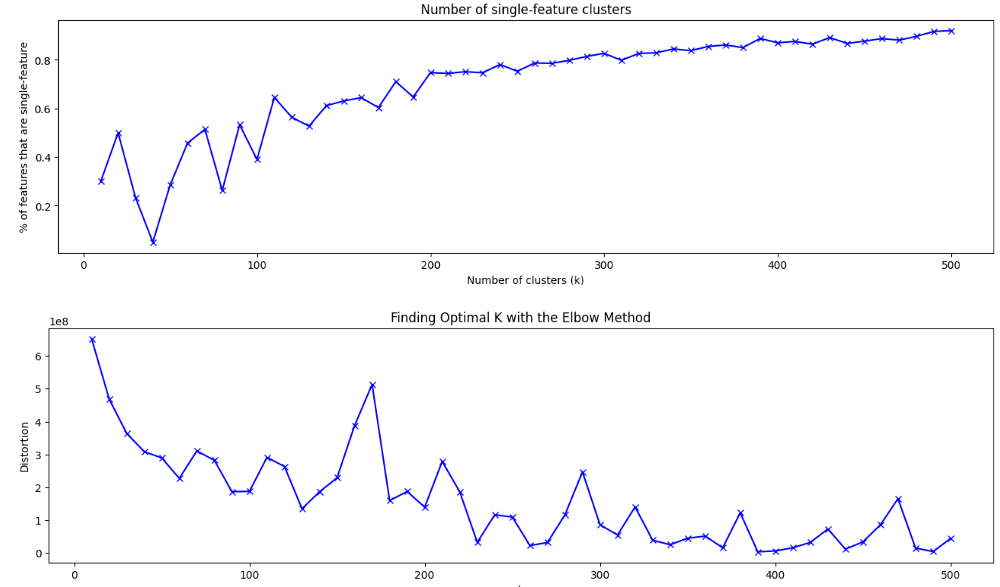}
\caption{Finding Optimum Number of Clusters \textmd{The top plot shows a plot of the number of single-feature clusters as the number of clusters increases. The bottom plot shows the sum of squared distances between each feature and the centroid of its clusters, obtained using the Elbow method. Both of these plots are used to determine the optimum number of clusters for this analysis.}}
\label{fig:biclustering_optimum}
\end{figure}

\ifCLASSOPTIONcaptionsoff
  \newpage
\fi


\begin{thebibliography}{10}
\providecommand{\url}[1]{#1}
\csname url@samestyle\endcsname
\providecommand{\newblock}{\relax}
\providecommand{\bibinfo}[2]{#2}
\providecommand{\BIBentrySTDinterwordspacing}{\spaceskip=0pt\relax}
\providecommand{\BIBentryALTinterwordstretchfactor}{4}
\providecommand{\BIBentryALTinterwordspacing}{\spaceskip=\fontdimen2\font plus
\BIBentryALTinterwordstretchfactor\fontdimen3\font minus
  \fontdimen4\font\relax}
\providecommand{\BIBforeignlanguage}[2]{{%
\expandafter\ifx\csname l@#1\endcsname\relax
\typeout{** WARNING: IEEEtran.bst: No hyphenation pattern has been}%
\typeout{** loaded for the language `#1'. Using the pattern for}%
\typeout{** the default language instead.}%
\else
\language=\csname l@#1\endcsname
\fi
#2}}
\providecommand{\BIBdecl}{\relax}
\BIBdecl


\bibitem{RN311}
\BIBentryALTinterwordspacing
S.~G. Priori, C.~Blomström-Lundqvist, A.~Mazzanti, N.~Blom, M.~Borggrefe,
  J.~Camm, P.~M. Elliott, D.~Fitzsimons, R.~Hatala, G.~Hindricks, P.~Kirchhof,
  K.~Kjeldsen, K.-H. Kuck, A.~Hernandez-Madrid, N.~Nikolaou, T.~M. Norekvål,
  C.~Spaulding, D.~J. Van~Veldhuisen, and E.~S.~D. Group, ``2015 esc guidelines
  for the management of patients with ventricular arrhythmias and the
  prevention of sudden cardiac death: The task force for the management of
  patients with ventricular arrhythmias and the prevention of sudden cardiac
  death of the european society of cardiology (esc)endorsed by: Association for
  european paediatric and congenital cardiology (aepc),'' \emph{European Heart
  Journal}, vol.~36, no.~41, pp. 2793--2867, 2015. [Online]. Available:
  \url{https://doi.org/10.1093/eurheartj/ehv316}
\BIBentrySTDinterwordspacing

\bibitem{RN355}
R.~J. Myerburg, ``Scientific gaps in the prediction and prevention of sudden
  cardiac death,'' \emph{J Cardiovasc Electrophysiol}, vol.~13, no.~7, pp.
  709--23, 2002.

\bibitem{RN358}
\BIBentryALTinterwordspacing
T.~Skjelbred, D.~Rajan, J.~Svane, T.~H. Lynge, and J.~Tfelt-Hansen, ``Sex
  differences in sudden cardiac death in a nationwide study of 54 028 deaths,''
  \emph{Heart}, vol. 108, no.~13, pp. 1012--1018, 2022. [Online]. Available:
  \url{https://heart.bmj.com/content/heartjnl/108/13/1012.full.pdf}
\BIBentrySTDinterwordspacing

\bibitem{RN361}
\BIBentryALTinterwordspacing
F.~N. Ågesen, T.~H. Lynge, P.~Blanche, J.~Banner, E.~Prescott, R.~Jabbari, and
  J.~Tfelt-Hansen, ``Temporal trends and sex differences in sudden cardiac
  death in the copenhagen city heart study,'' \emph{Heart}, vol. 107, no.~16,
  pp. 1303--1309, 2021. [Online]. Available:
  \url{https://heart.bmj.com/content/heartjnl/107/16/1303.full.pdf}
\BIBentrySTDinterwordspacing

\bibitem{RN369}
M.~Hayashi, W.~Shimizu, and C.~M. Albert, ``The spectrum of epidemiology
  underlying sudden cardiac death,'' \emph{Circ Res}, vol. 116, no.~12, pp.
  1887--906, 2015.

\bibitem{RN291}
\BIBentryALTinterwordspacing
J.~M. Robert and M.~J. Junttila, ``Sudden cardiac death caused by coronary
  heart disease,'' \emph{Circulation}, vol. 125, no.~8, pp. 1043--1052, 2012.
  [Online]. Available:
  \url{https://www.ahajournals.org/doi/abs/10.1161/CIRCULATIONAHA.111.023846
  ,eprint =
  https://www.ahajournals.org/doi/pdf/10.1161/CIRCULATIONAHA.111.023846}
\BIBentrySTDinterwordspacing

\bibitem{RN297}
\BIBentryALTinterwordspacing
R.~Deo and C.~M. Albert, ``Epidemiology and genetics of sudden cardiac death,''
  \emph{Circulation}, vol. 125, no.~4, pp. 620--637, 2012, doi:
  10.1161/CIRCULATIONAHA.111.023838. [Online]. Available:
  \url{https://doi.org/10.1161/CIRCULATIONAHA.111.023838}
\BIBentrySTDinterwordspacing

\bibitem{RN294}
\BIBentryALTinterwordspacing
J.~M. Jacqueline, A.~P.~M. Gorgels, W.~I. Dubois-Arbouw, J.~W., M.~J. A.~P.
  Daemen, L.~G.~E. Houben, and H.~J.~J. Wellens, ``Out-of-hospital cardiac
  arrest in the 1990s: A population-based study in the maastricht area on
  incidence, characteristics and survival,'' \emph{Journal of the American
  College of Cardiology}, vol.~30, no.~6, pp. 1500--1505, 1997. [Online].
  Available:
  \url{https://www.sciencedirect.com/science/article/pii/S0735109797003550}
\BIBentrySTDinterwordspacing

\bibitem{RN277}
\BIBentryALTinterwordspacing
A.~A.~H. de~Hond, A.~M. Leeuwenberg, L.~Hooft, I.~M.~J. Kant, S.~W.~J. Nijman,
  H.~J.~A. van Os, J.~J. Aardoom, T.~P.~A. Debray, E.~Schuit, M.~van Smeden,
  J.~B. Reitsma, E.~W. Steyerberg, N.~H. Chavannes, and K.~G.~M. Moons,
  ``Guidelines and quality criteria for artificial intelligence-based
  prediction models in healthcare: a scoping review,'' \emph{npj Digital
  Medicine}, vol.~5, no.~1, p.~2, 2022. [Online]. Available:
  \url{https://doi.org/10.1038/s41746-021-00549-7}
\BIBentrySTDinterwordspacing

\bibitem{10.1093/eurheartj/ehu207}
\BIBentryALTinterwordspacing
E.~W. Steyerberg and Y.~Vergouwe, ``{Towards better clinical prediction models:
  seven steps for development and an ABCD for validation},'' \emph{European
  Heart Journal}, vol.~35, no.~29, pp. 1925--1931, 06 2014. [Online].
  Available: \url{https://doi.org/10.1093/eurheartj/ehu207}
\BIBentrySTDinterwordspacing

\bibitem{10.1093/eurheartj/ehac238}
\BIBentryALTinterwordspacing
M.~van Smeden, G.~Heinze, B.~Van~Calster, F.~W. Asselbergs, P.~E. Vardas,
  N.~Bruining, P.~de~Jaegere, J.~H. Moore, S.~Denaxas, A.~L. Boulesteix, and
  K.~G.~M. Moons, ``{Critical appraisal of artificial intelligence-based
  prediction models for cardiovascular disease },'' \emph{European Heart
  Journal}, vol.~43, no.~31, pp. 2921--2930, 05 2022. [Online]. Available:
  \url{https://doi.org/10.1093/eurheartj/ehac238}
\BIBentrySTDinterwordspacing

\bibitem{RN192}
\BIBentryALTinterwordspacing
C.~Rudin, ``Stop explaining black box machine learning models for high stakes
  decisions and use interpretable models instead,'' \emph{Nature Machine
  Intelligence}, vol.~1, no.~5, pp. 206--215, 2019. [Online]. Available:
  \url{https://doi.org/10.1038/s42256-019-0048-x}
\BIBentrySTDinterwordspacing

\bibitem{RN195}
\BIBentryALTinterwordspacing
M.~E. Fenech and O.~Buston, ``Ai in cardiac imaging: A uk-based perspective on
  addressing the ethical, social, and political challenges,'' \emph{Frontiers
  in cardiovascular medicine}, vol.~7, pp. 54--54, 2020. [Online]. Available:
  \url{https://pubmed.ncbi.nlm.nih.gov/32351974}
\BIBentrySTDinterwordspacing

\bibitem{RN196}
\BIBentryALTinterwordspacing
E.~M. Cahan, T.~Hernandez-Boussard, S.~Thadaney-Israni, and D.~L. Rubin,
  ``Putting the data before the algorithm in big data addressing personalized
  healthcare,'' \emph{npj Digital Medicine}, vol.~2, no.~1, p.~78, 2019.
  [Online]. Available: \url{https://doi.org/10.1038/s41746-019-0157-2}
\BIBentrySTDinterwordspacing

\bibitem{RN197}
\BIBentryALTinterwordspacing
K.~Harron, A.~Wade, R.~Gilbert, B.~Muller-Pebody, and H.~Goldstein,
  ``Evaluating bias due to data linkage error in electronic healthcare
  records,'' \emph{BMC Medical Research Methodology}, vol.~14, no.~1, p.~36,
  2014. [Online]. Available: \url{https://doi.org/10.1186/1471-2288-14-36}
\BIBentrySTDinterwordspacing

\bibitem{RN207}
\BIBentryALTinterwordspacing
A.~J. Larrazabal, N.~Nieto, V.~Peterson, D.~H. Milone, and E.~Ferrante,
  ``Gender imbalance in medical imaging datasets produces biased classifiers
  for computer-aided diagnosis,'' \emph{Proceedings of the National Academy of
  Sciences}, vol. 117, no.~23, pp. 12\,592--12\,594, 2020. [Online]. Available:
  \url{https://www.pnas.org/content/pnas/117/23/12592.full.pdf}
\BIBentrySTDinterwordspacing

\bibitem{RN208}
\BIBentryALTinterwordspacing
I.~Y. Chen, E.~Pierson, S.~Rose, S.~Joshi, K.~Ferryman, and M.~Ghassemi,
  ``Ethical machine learning in healthcare,'' \emph{Annual Review of Biomedical
  Data Science}, vol.~4, no.~1, pp. 123--144, 2021. [Online]. Available:
  \url{https://www.annualreviews.org/doi/abs/10.1146/annurev-biodatasci-092820-114757}
\BIBentrySTDinterwordspacing

\bibitem{RN261}
\BIBentryALTinterwordspacing
Y.~Li, S.~Rao, J.~R.~A. Solares, A.~Hassaine, R.~Ramakrishnan, D.~Canoy,
  Y.~Zhu, K.~Rahimi, and G.~Salimi-Khorshidi, ``Behrt: Transformer for
  electronic health records,'' \emph{Scientific Reports}, vol.~10, no.~1, p.
  7155, 2020. [Online]. Available:
  \url{https://doi.org/10.1038/s41598-020-62922-y}
\BIBentrySTDinterwordspacing

\bibitem{RN264}
P.~Nguyen, T.~Truyen, W.~Nilmini, and V.~Svetha, ``Deepr: A convolutional net
  for medical records,'' \emph{IEEE Journal of Biomedical and Health
  Informatics}, vol.~21, no.~1, pp. 22--30, 2017.

\bibitem{RN265}
\BIBentryALTinterwordspacing
R.~Miotto, L.~Li, B.~A. Kidd, and J.~T. Dudley, ``Deep patient: An unsupervised
  representation to predict the future of patients from the electronic health
  records,'' \emph{Scientific Reports}, vol.~6, no.~1, p. 26094, 2016.
  [Online]. Available: \url{https://doi.org/10.1038/srep26094}
\BIBentrySTDinterwordspacing

\bibitem{9706318}
S.~Rao, Y.~Li, R.~Ramakrishnan, A.~Hassaine, D.~Canoy, J.~Cleland,
  T.~Lukasiewicz, G.~Salimi-Khorshidi, and K.~Rahimi, ``An explainable
  transformer-based deep learning model for the prediction of incident heart
  failure,'' \emph{IEEE Journal of Biomedical and Health Informatics}, vol.~26,
  no.~7, pp. 3362--3372, 2022.

\bibitem{RN262}
\BIBentryALTinterwordspacing
J.~Roberto, F.~Elisa, Y.~Zhu, F.~Rahimian, D.~Canoy, J.~Tran, A.~Catarina,
  H.~P. Amir, M.~Zottoli, M.~Nazarzadeh, N.~Conrad, K.~Rahimi, and
  G.~Salimi-Khorshidi, ``Deep learning for electronic health records: A
  comparative review of multiple deep neural architectures,'' \emph{Journal of
  Biomedical Informatics}, vol. 101, p. 103337, 2020. [Online]. Available:
  \url{https://www.sciencedirect.com/science/article/pii/S1532046419302564}
\BIBentrySTDinterwordspacing

\bibitem{BERT_ref}
\BIBentryALTinterwordspacing
J.~Devlin, M.-W. Chang, K.~Lee, and K.~Toutanova, ``Bert: Pre-training of deep
  bidirectional transformers for language understanding,'' \emph{arXiv
  preprint}, 2018. [Online]. Available: \url{https://arxiv.org/abs/1810.04805}
\BIBentrySTDinterwordspacing

\bibitem{RN412}
M.~Fenech and B.~Oliver, ``Ai in cardiac imaging: A uk-based perspective on
  addressing the ethical, social, and political challenges,'' \emph{Frontiers
  in Cardiovascular Medicine}, vol.~7, 2020.

\bibitem{RN414}
\BIBentryALTinterwordspacing
W.~J. Murdoch, C.~Singh, K.~Kumbier, R.~Abbasi-Asl, and B.~Yu, ``Definitions,
  methods, and applications in interpretable machine learning,''
  \emph{Proceedings of the National Academy of Sciences}, vol. 116, no.~44, pp.
  22\,071--22\,080, 2019. [Online]. Available:
  \url{https://www.pnas.org/doi/abs/10.1073/pnas.1900654116 ,eprint =
  https://www.pnas.org/doi/pdf/10.1073/pnas.1900654116}
\BIBentrySTDinterwordspacing

\bibitem{RN413}
\BIBentryALTinterwordspacing
S.~Vollmer, B.~A. Mateen, G.~Bohner, F.~J. Király, R.~Ghani, P.~Jonsson,
  S.~Cumbers, A.~Jonas, K.~S.~L. McAllister, P.~Myles, D.~Grainger, M.~Birse,
  R.~Branson, K.~G.~M. Moons, G.~S. Collins, J.~P.~A. Ioannidis, C.~Holmes, and
  H.~Hemingway, ``Machine learning and artificial intelligence research for
  patient benefit: 20 critical questions on transparency, replicability,
  ethics, and effectiveness,'' \emph{BMJ}, vol. 368, p. l6927, 2020. [Online].
  Available: \url{http://www.bmj.com/content/368/bmj.l6927.abstract}
\BIBentrySTDinterwordspacing

\bibitem{RN263}
J.~Vig, ``A multiscale visualization of attention in the transformer model,''
  pp. 37--42, 01 2019.

\bibitem{biclustering_genes}
\BIBentryALTinterwordspacing
H.~Aouabed, R.~Santamaria, and M.~Elloumi, ``Visualizing biclustering results
  on gene expression data: A survey,'' in \emph{2021 13th International
  Conference on Bioinformatics and Biomedical Technology}, ser. ICBBT
  2021.\hskip 1em plus 0.5em minus 0.4em\relax New York, NY, USA: Association
  for Computing Machinery, 2021, p. 170–179. [Online]. Available:
  \url{https://doi.org/10.1145/3473258.3473284}
\BIBentrySTDinterwordspacing

\bibitem{RN539}
\BIBentryALTinterwordspacing
F.~Divina, F.~A. Goméz~Vela, and M.~García~Torres, ``Biclustering of smart
  building electric energy consumption data,'' \emph{Applied Sciences}, vol.~9,
  no.~2, 2019. [Online]. Available:
  \url{https://www.mdpi.com/2076-3417/9/2/222}
\BIBentrySTDinterwordspacing

\bibitem{Dhamodharavadhani2021}
\BIBentryALTinterwordspacing
S.~Dhamodharavadhani and R.~Rathipriya, \emph{Biclustering Analysis of
  Countries Using COVID-19 Epidemiological Data}.\hskip 1em plus 0.5em minus
  0.4em\relax Cham: Springer International Publishing, 2021, pp. 93--114.
  [Online]. Available: \url{https://doi.org/10.1007/978-3-030-70478-0\_6}
\BIBentrySTDinterwordspacing

\bibitem{RN100}
\BIBentryALTinterwordspacing
``Glasgow safehaven data catalogue,'' 2020. [Online]. Available:
  \url{https://www.nhsggc.org.uk/media/266676/glasgow-safe-haven-dataset-catalogue.pdf}
\BIBentrySTDinterwordspacing

\bibitem{RN247}
E.~Choi, B.~M. Taha, K.~J. A., S.~Andy, S.~W. F., and S.~Jimeng, ``Retain: An
  interpretable predictive model for healthcare using reverse time attention
  mechanism,'' pp. 3512–3520 , numpages = 9, 2016.

\bibitem{RN249}
\BIBentryALTinterwordspacing
T.~Tran, T.~D. Nguyen, D.~Phung, and V.~Svetha, ``Learning vector
  representation of medical objects via emr-driven nonnegative restricted
  boltzmann machines (enrbm),'' \emph{Journal of Biomedical Informatics},
  vol.~54, pp. 96--105, 2015. [Online]. Available:
  \url{https://www.sciencedirect.com/science/article/pii/S1532046415000143}
\BIBentrySTDinterwordspacing

\bibitem{coclustering_original}
\BIBentryALTinterwordspacing
I.~S. Dhillon, ``Co-clustering documents and words using bipartite spectral
  graph partitioning,'' in \emph{Proceedings of the Seventh ACM SIGKDD
  International Conference on Knowledge Discovery and Data Mining}, ser. KDD
  '01.\hskip 1em plus 0.5em minus 0.4em\relax New York, NY, USA: Association
  for Computing Machinery, 2001, p. 269–274. [Online]. Available:
  \url{https://doi.org/10.1145/502512.502550}
\BIBentrySTDinterwordspacing

\bibitem{RN550}
Y.~Li, M.~Mohammad, S.-K. Gholamreza, R.~Shishir, H.~Abdelaali, C.~Dexter,
  L.~Thomas, and R.~Kazem, ``Hi-behrt: Hierarchical transformer-based model for
  accurate prediction of clinical events using multimodal longitudinal
  electronic health records,'' 2021.

\bibitem{RN377}
S.~Angraal, B.~J. Mortazavi, A.~Gupta, R.~Khera, T.~Ahmad, N.~R. Desai, D.~L.
  Jacoby, F.~A. Masoudi, J.~A. Spertus, and H.~M. Krumholz, ``Machine learning
  prediction of mortality and hospitalization in heart failure with preserved
  ejection fraction,'' \emph{JACC Heart Fail}, vol.~8, no.~1, pp. 12--21, 2020.

\bibitem{RN558}
\BIBentryALTinterwordspacing
R.~J. Desai, S.~V. Wang, M.~Vaduganathan, T.~Evers, and S.~Schneeweiss,
  ``Comparison of machine learning methods with traditional models for use of
  administrative claims with electronic medical records to predict heart
  failure outcomes,'' \emph{JAMA Network Open}, vol.~3, no.~1, pp.
  e1\,918\,962--e1\,918\,962, 2020. [Online]. Available:
  \url{https://doi.org/10.1001/jamanetworkopen.2019.18962}
\BIBentrySTDinterwordspacing

\bibitem{RN560}
M.~Tokodi, W.~R. Schwertner, A.~Kovács, Z.~Tősér, L.~Staub, A.~Sárkány,
  B.~K. Lakatos, A.~Behon, A.~M. Boros, P.~Perge, V.~Kutyifa, G.~Széplaki,
  L.~Gellér, B.~Merkely, and A.~Kosztin, ``Machine learning-based mortality
  prediction of patients undergoing cardiac resynchronization therapy: the
  semmelweis-crt score,'' no. 1522-9645 (Electronic).

\bibitem{RN561}
\BIBentryALTinterwordspacing
W.~Qiu, H.~Chen, A.~B. Dincer, S.~Lundberg, M.~Kaeberlein, and S.-I. Lee,
  ``Interpretable machine learning prediction of all-cause mortality,''
  \emph{Communications Medicine}, vol.~2, no.~1, p. 125, 2022. [Online].
  Available: \url{https://doi.org/10.1038/s43856-022-00180-x}
\BIBentrySTDinterwordspacing

\bibitem{RN563}
\BIBentryALTinterwordspacing
O.~P. van~der Galiën, R.~C. Hoekstra, M.~T. Gürgöze, O.~C. Manintveld, M.~R.
  van~den Bunt, C.~J. Veenman, and E.~Boersma, ``Prediction of long-term
  hospitalisation and all-cause mortality in patients with chronic heart
  failure on dutch claims data: a machine learning approach,'' \emph{BMC
  Medical Informatics and Decision Making}, vol.~21, no.~1, p. 303, 2021.
  [Online]. Available: \url{https://doi.org/10.1186/s12911-021-01657-w}
\BIBentrySTDinterwordspacing

\bibitem{RN538}
\BIBentryALTinterwordspacing
Y.~Li, G.~Salimi-Khorshidi, S.~Rao, D.~Canoy, A.~Hassaine, T.~Lukasiewicz,
  K.~Rahimi, and M.~Mamouei, ``Validation of risk prediction models applied to
  longitudinal electronic health record data for the prediction of major
  cardiovascular events in the presence of data shifts,'' \emph{European Heart
  Journal - Digital Health}, vol.~3, no.~4, pp. 535--547, 2022. [Online].
  Available: \url{https://doi.org/10.1093/ehjdh/ztac061}
\BIBentrySTDinterwordspacing

\bibitem{RN553}
\BIBentryALTinterwordspacing
L.~Rasmy, Y.~Xiang, Z.~Xie, C.~Tao, and D.~Zhi, ``Med-bert: pretrained
  contextualized embeddings on large-scale structured electronic health records
  for disease prediction,'' \emph{npj Digital Medicine}, vol.~4, no.~1, p.~86,
  2021. [Online]. Available: \url{https://doi.org/10.1038/s41746-021-00455-y}
\BIBentrySTDinterwordspacing

\bibitem{RN548}
J.~Shang, M.~Tengfei, X.~Cao, and S.~J., ``Pre-training of graph augmented
  transformers for medication recommendation,'' pp. 5953--5959, 08 2019.

\bibitem{RN551}
\BIBentryALTinterwordspacing
C.~Pang, X.~Jiang, K.~S. Kalluri, M.~Spotnitz, R.~Chen, A.~Perotte, and
  K.~Natarajan, ``Cehr-bert: Incorporating temporal information from structured
  ehr data to improve prediction tasks,'' pp. 239--260, 2021. [Online].
  Available: \url{https://proceedings.mlr.press/v158/pang21a.html}
\BIBentrySTDinterwordspacing

\bibitem{RN549}
S.~Rao, L.~Y., R.~Rema, H.~Abdelâali, C.~Dexter, Z.~Yajie, S.-K. G., and
  R.~Kazem, ``Behrt-hf: an interpretable transformer-based, deep learning model
  for prediction of incident heart failure,'' \emph{European Heart Journal},
  vol.~41, 2020.

\bibitem{RN562}
\BIBentryALTinterwordspacing
K.~Wang, J.~Tian, C.~Zheng, H.~Yang, J.~Ren, Y.~Liu, Q.~Han, and Z.~Yanbo,
  ``Interpretable prediction of 3-year all-cause mortality in patients with
  heart failure caused by coronary heart disease based on machine learning and
  shap,'' \emph{Computers in Biology and Medicine}, vol. 137, p. 104813, 2021.
  [Online]. Available:
  \url{https://www.sciencedirect.com/science/article/pii/S0010482521006077}
\BIBentrySTDinterwordspacing

\bibitem{RN326}
\BIBentryALTinterwordspacing
J.~Hewitt, M.~Walters, S.~Padmanabhan, and J.~Dawson, ``Cohort profile of the
  uk biobank: diagnosis and characteristics of cerebrovascular disease,''
  \emph{BMJ Open}, vol.~6, no.~3, p. e009161, 2016. [Online]. Available:
  \url{http://bmjopen.bmj.com/content/6/3/e009161.abstract}
\BIBentrySTDinterwordspacing

\bibitem{RN327}
A.~Fry, T.~J. Littlejohns, C.~Sudlow, N.~Doherty, L.~Adamska, T.~Sprosen,
  R.~Collins, and N.~E. Allen, ``Comparison of sociodemographic and
  health-related characteristics of uk biobank participants with those of the
  general population,'' \emph{Am J Epidemiol}, vol. 186, no.~9, pp. 1026--1034,
  2017.

\bibitem{RN424}
K.~Clark, K.~Urvashi, L.~Omer, and M.~C. D., ``What does bert look at? an
  analysis of bert's attention,'' \emph{arXiv preprint arXiv:1906.04341}, 2019.

\bibitem{RN421}
S.~Jain and C.~W. Byron, ``Attention is not explanation,'' 2019.

\bibitem{RN194}
\BIBentryALTinterwordspacing
H.~Kaur, N.~Harsha, J.~Samuel, C.~Rich, W.~Hanna, and W.~V. Jennifer,
  ``Interpreting interpretability: Understanding data scientists' use of
  interpretability tools for machine learning,'' 2020. [Online]. Available:
  \url{https://doi.org/10.1145/3313831.3376219}
\BIBentrySTDinterwordspacing

\bibitem{RN203}
J.~Dieber and K.~Sabrina, ``Why model why? assessing the strengths and
  limitations of lime,'' 2020.

\bibitem{RN210}
B.~Kim, K.~Rajiv, and K.~Oluwasanmi, ``Examples are not enough, learn to
  criticize! criticism for interpretability,'' pp. 2288–2296 , numpages = 9,
  2016.

\bibitem{RN522}
S.~Rao, L.~Yikuan, R.~Rema, H.~Abdelaali, C.~Dexter, C.~John, L.~Thomas, S.-K.
  Gholamreza, and R.~Kazem, ``An explainable transformer-based deep learning
  model for the prediction of incident heart failure,'' \emph{IEEE Journal of
  Biomedical and Health Informatics}, vol.~26, no.~7, pp. 3362--3372, 2022.

\bibitem{Jones_2020}
\BIBentryALTinterwordspacing
Y.~Jones, F.~Deligianni, and J.~Dalton, ``Improving ecg classification
  interpretability using saliency maps,'' in \emph{2020 {IEEE} 20th
  International Conference on Bioinformatics and Bioengineering
  ({BIBE})}.\hskip 1em plus 0.5em minus 0.4em\relax {IEEE}, oct 2020. [Online].
  Available: \url{https://doi.org/10.1109\%2Fbibe50027.2020.00114}
\BIBentrySTDinterwordspacing

\bibitem{NIPS2017_8a20a862}
\BIBentryALTinterwordspacing
S.~M. Lundberg and S.-I. Lee, ``A unified approach to interpreting model
  predictions,'' in \emph{Advances in Neural Information Processing Systems},
  I.~Guyon, U.~V. Luxburg, S.~Bengio, H.~Wallach, R.~Fergus, S.~Vishwanathan,
  and R.~Garnett, Eds., vol.~30.\hskip 1em plus 0.5em minus 0.4em\relax Curran
  Associates, Inc., 2017. [Online]. Available:
  \url{https://proceedings.neurips.cc/paper/2017/file/8a20a8621978632d76c43dfd28b67767-Paper.pdf}
\BIBentrySTDinterwordspacing

\bibitem{RN457}
C.~Morton, ``Health inequalities in scotland,'' 2020.

\bibitem{RN483}
D.~A. Fixsen, D.~S. Barrett, and M.~Shimonovich, ``Supporting vulnerable
  populations during the pandemic: Stakeholders' experiences and perceptions of
  social prescribing in scotland during covid-19,'' \emph{Qual Health Res},
  vol.~32, no.~4, pp. 670--682, 2022.

\bibitem{RN458}
\BIBentryALTinterwordspacing
``Understanding glasgow: The glasgow indictors project,'' 2021. [Online].
  Available:
  \url{https://www.understandingglasgow.com/indicators/poverty/overview}
\BIBentrySTDinterwordspacing

\bibitem{RN289}
Y.~Meng, S.~William, O.~Michael, and A.~C. W., ``Hcet: Hierarchical clinical
  embedding with topic modeling on electronic health records for predicting
  future depression,'' \emph{IEEE Journal of Biomedical and Health
  Informatics}, vol.~25, no.~4, pp. 1265--1272, 2021.

\bibitem{RN278}
\BIBentryALTinterwordspacing
W.~L. Cava, C.~Bauer, J.~H. Moore, and S.~A. Pendergrass, ``Interpretation of
  machine learning predictions for patient outcomes in electronic health
  records,'' \emph{AMIA ... Annual Symposium proceedings. AMIA Symposium}, vol.
  2019, pp. 572--581, 2020. [Online]. Available:
  \url{https://pubmed.ncbi.nlm.nih.gov/32308851}
\BIBentrySTDinterwordspacing

\bibitem{RN260}
\BIBentryALTinterwordspacing
M.~T. Ribeiro, S.~Sameer, and G.~Carlos, ``Why should i trust you? : Explaining
  the predictions of any classifier,'' pp. 1135–1144 , numpages = 10, 2016.
  [Online]. Available: \url{https://doi.org/10.1145/2939672.2939778}
\BIBentrySTDinterwordspacing

\bibitem{RN275}
\BIBentryALTinterwordspacing
W.~Webber, M.~Alistair, and Z.~Justin, ``A similarity measure for indefinite
  rankings,'' \emph{ACM Trans. Inf. Syst.}, vol.~28, no.~4, 2010. [Online].
  Available: \url{https://doi.org/10.1145/1852102.1852106}
\BIBentrySTDinterwordspacing

\end{thebibliography}
\end{document}